\documentclass{article}

\usepackage{arxiv}

\usepackage[utf8]{inputenc} % allow utf-8 input
\usepackage[T1]{fontenc}    % use 8-bit T1 fonts
\usepackage{hyperref}       % hyperlinks
\usepackage{url}            % simple URL typesetting
\usepackage{booktabs}       % professional-quality tables
\usepackage{amsfonts}       % blackboard math symbols
\usepackage{microtype}      % microtypography
\usepackage{lipsum}
\usepackage{graphicx}
\graphicspath{ {./images/} }

\usepackage{subcaption} % Allows for subfigures
\usepackage{xspace}
\usepackage{amsmath}
\usepackage{siunitx}

\DeclareMathOperator*{\argmin}{argmin} % no space, limits underneath in displays

\title{Federated Learning from Molecules to Processes:\\ A Perspective}

\author{
  Jan G. Rittig \\
  Process Systems Engineering (AVT.SVT)\\
  RWTH Aachen University\\
  \texttt{jan.rittig@rwth-aachen.de} \\
  \textit{Corresponding author}
   \And
  Clemens Kortmann \\
  Process Systems Engineering (AVT.SVT)\\
  RWTH Aachen University\\
  \texttt{clemens.kortmann@rwth-aachen.de} \\
}

\begin{document}

\newcommand{\ChemicalEngineering}{ChemE\xspace}
\newcommand{\MachineLearning}{ML\xspace}
\newcommand{\FederatedLearning}{FL\xspace}
	
\maketitle
\begin{abstract}

We present a perspective on federated learning in chemical engineering that envisions collaborative efforts in machine learning (ML) developments within the chemical industry.
Large amounts of chemical and process data are proprietary to chemical companies and are therefore locked in data silos, hindering the training of ML models on large data sets in chemical engineering.
Recently, the concept of federated learning has gained increasing attention in ML research, enabling organizations to jointly train machine learning models without disclosure of their individual data.
We discuss potential applications of federated learning in several fields of chemical engineering, from the molecular to the process scale.
In addition, we apply federated learning in two exemplary case studies that simulate practical scenarios of multiple chemical companies holding proprietary data sets: (i) prediction of binary mixture activity coefficients with graph neural networks and (ii) system identification of a distillation column with autoencoders.
Our results indicate that ML models jointly trained with federated learning yield significantly higher accuracy than models trained by each chemical company individually and can perform similarly to models trained on combined datasets from all companies.
Federated learning has therefore great potential to advance ML models in chemical engineering while respecting corporate data privacy, making it promising for future industrial applications.

\end{abstract}
\section{Introduction}\label{sec:Intro}

\noindent Machine learning (\MachineLearning) methods have led to remarkable progress in diverse fields of chemical engineering (\ChemicalEngineering) research~\cite{Schweidtmann.2021, StriethKalthoff.2020, Daoutidis.2024, Cheng.2024}, but data scarcity remains a key limitation. 
Scaling up data set sizes has been a key factor for recent advances in \MachineLearning applications such as natural language processing and computer vision, cf. neural scaling laws~\cite{hestness2017deep, kaplan2020scaling, hoffmann2022training, zhai2022scaling}.
While large data sets of texts and images with billions of data points can be collected from the internet, \ChemicalEngineering applications typically face smaller data sets of hundreds to thousands of experimental data points and data sets with low variance, e.g., chemical process data at similar operating points, cf.~\cite{qin2019advances, thebelt2022maximizing}. 
The vast majority of this highly valuable chemical and process data has been collected by chemical enterprises over the last decades, and is therefore locked in their data silos~\cite{Zhu.2022}.
In fact, the companies have a strong interest in keeping their data private because they have invested a substantial amount of money and effort in obtaining it, and because it contains confidential information about the chemical space and the processes they are working on~\cite{Dutta.2024}.
As a result, only a few datasets are publicly available for \ChemicalEngineering applications, with notable exceptions in molecular property prediction~\cite{geballe2010sampl2, mobley2014freesolv, qm9}, reaction prediction and optimization~\cite{felton2021summit, wigh2024orderly}, and process fault detection~\cite{chiang2012fault}.
The generally low public availability of large chemical and process datasets, and consequently missing benchmarks, limits \MachineLearning research and the development of large-scale models for \ChemicalEngineering.

Several approaches have been considered to mitigate the problem of scarce chemical data.
These include the incorporation of domain knowledge into the model, e.g., in hybrid architectures~\cite{bradley2022perspectives, schweidtmann2024review} such as physics-informed \MachineLearning~\cite{karniadakis2021physics} and agentic architectures coupled with domain-specific tools~\cite{boiko2023autonomous, m2024augmenting}, which can reduce the amount of data needed for model training.
Another way is to artificially increase the amount of data available for model training through approaches like multi-task, multi-fidelity, self-supervised, and transfer learning.
For example, in transfer learning, data from related \ChemicalEngineering tasks is combined, whereas in multi-fidelity learning, high-fidelity experimental data is augmented with low-fidelity data from simulations, see, e.g.,~\cite{vermeire2021transfer, wu2020fault, li2020transfer, savage2023multi, nevolianis2024multi}.
Although these approaches have been shown to increase the performance of ML models, many \MachineLearning applications in \ChemicalEngineering still lack sufficient experimental data for training or rely solely on simulated data.

We argue that leveraging private chemical and process data sets to increase the quantity and quality of the available data used for model training will be crucial to catalyze \MachineLearning for \ChemicalEngineering applications.
Therefore, an approach is needed that (a) leverages existing data from chemical enterprises without harming their data privacy and (b) enables new kinds of research cooperation between multiple industrial and possibly academic partners to value unexplored chemical data on a larger scale.
To this end, federated learning (\FederatedLearning) is particularly promising.

\begin{figure}[h!bt]
	\begin{center}
		\includegraphics[width=\textwidth, keepaspectratio]{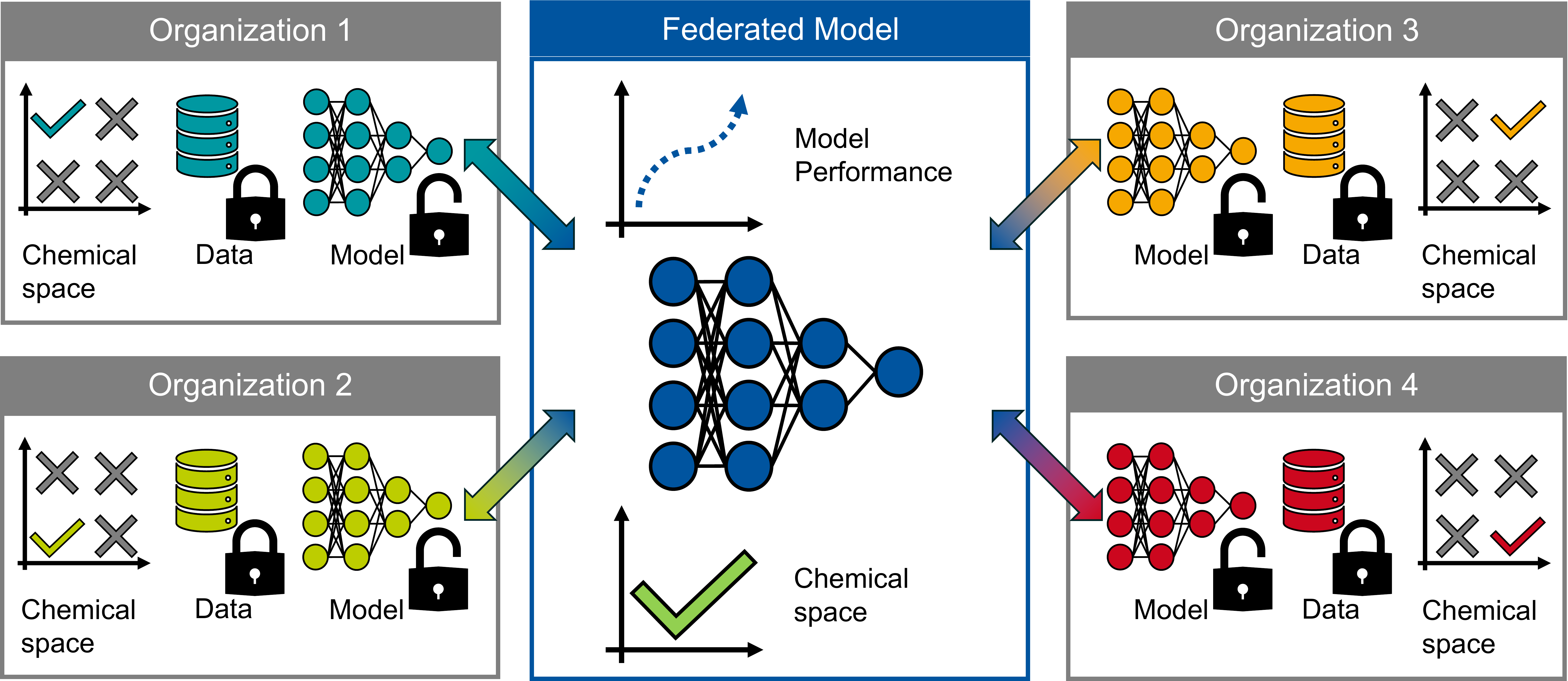}
		\caption{Schematic illustration of four chemical companies working on different regions in the chemical space and using federated learning to jointly advance a machine learning model while respecting data privacy.}
		\label{fig:FL_concept}
	\end{center}
\end{figure}

\FederatedLearning enables multiple entities to jointly train \MachineLearning models without disclosure of their individual data~\cite{McMahan.2017, Kairouz.2019, Daly.2024}.
The \FederatedLearning approach has gained increasing attention in recent years for applications of \MachineLearning where data privacy is a main concern~\cite{Liu.2024a, Wen.2023}.
Figure~\ref{fig:FL_concept} illustrates the concept of \FederatedLearning for the example of four chemical companies that have private data on different regions of the chemical space and jointly train a neural network.
Specifically, each company trains a \MachineLearning model on its local data and only exchanges the trained model parameters, so that the individual models can be aggregated into a global model, which is ideally more accurate and has a wider applicability range than the individual ones.
Hence, private data can be utilized for collaborative model development without leaving the individual entities.

\FederatedLearning has been utilized by both academia and industry in various domains, including natural language processing~\cite{Li.2024}, finance~\cite{Wen.2023}, and healthcare~\cite{rieke2020future, Nguyen.2023b, Rauniyar.2024, Heyndrickx.2024}. 
One of the first \FederatedLearning applications points to Google setting up a federation of millions of individual mobile phone users to train recurrent neural networks for next-word prediction in text messages on Google keyboards -- without the need that users share private text message data~\cite{yang2018applied}.
Another example from natural language processing is the use of \FederatedLearning to pre-train billion-scale large language models (LLMs)~\cite{Sani.2024a}.
Furthermore, \FederatedLearning has been applied in healthcare for training convolutional neural networks in detecting cancer from medical images based on patient data locally stored in numerous healthcare institutions~\cite{lu2022federated, guan2024federated}.
On an industry scale, the MELLODDY project serves as a prime \FederatedLearning project~\cite{Heyndrickx.2023}: ten pharmaceutical companies teamed up with academia to develop quantitative structure-activity relationships (QSARs) in collaborative efforts.
Specifically, the companies jointly trained neural networks with extended connectivity fingerprints as input for drug property prediction, demonstrating improved prediction accuracy and applicability range compared to the models available at the individual companies~\cite{Heyndrickx.2024, Heyndrickx.2023, Oldenhof.2023}. 
More recently, Hanser et al. demonstrated improved bioactivity prediction capabilities of ML models by combining \FederatedLearning with knowledge distillation (cf.~\cite{wu2022communication}) in a collaboration with eight pharmaceutical companies~\cite{Hanser.2025}.
Overall, these examples showcase that \FederatedLearning has already demonstrated great potential in many domains and practical applications.

Given the promising results achieved with \FederatedLearning in various domains, we study \FederatedLearning as an approach to address data scarcity and advance \MachineLearning in \ChemicalEngineering. 

Reviewing existing \FederatedLearning works in \ChemicalEngineering engineering, we only find a few recent applications:
Hayer investigated \FederatedLearning to jointly train neural networks to predict process variables of aluminum electrolysis based on process data stored at different chemical producers, showing slight improvements in predictive accuracy~\cite{Hayer.2021}. 
Furthermore, Zhu et al.~\cite{Zhu.2022} developed \FederatedLearning approaches to train graph neural networks for molecular property prediction and propose FedChem, a benchmark for predicting a variety of physicochemical and biological properties when data is heterogeneously distributed across different entities.
Recently, Xu~\&~Wu~\cite{Xu.2024} proposed \FederatedLearning of a data-driven model for distributed model predictive control of two chemical reactors in series without the need to exchange data between the two reactor models, thereby increasing resilience against cyber attacks.
Notably, Dutta et al.~\cite{Dutta.2024} provide a tutorial of \FederatedLearning in relation to \ChemicalEngineering and present pharmaceutical examples.
Applications of \FederatedLearning also emerge in other domains strongly related to \ChemicalEngineering, e.g., in energy systems for power production forecasting~\cite{Cheng.2022, grataloup2024review}.
In general, \FederatedLearning for \ChemicalEngineering is recently emerging but is largely underexplored.

We here present a perspective on \FederatedLearning in \ChemicalEngineering. 
Specifically, we first describe the concept of \FederatedLearning (Section~\ref{text:sota}).
We then discuss the aspects of \FederatedLearning in the context of \ChemicalEngineering and highlight potential applications, such as flowsheet digitization and process design (Section~\ref{text:FLCE}).
Then, we present two exemplary case studies: 
We propose \FederatedLearning for prediction of infinite dilution activity coefficients of binary mixtures with graph neural networks (Section~\ref{text:gnn}) and for system identification of a dynamic chemical process using autoencoders (Section~\ref{text:mpc}).
Both case studies show that collaboration of chemical companies through \FederatedLearning can increase predictive accuracies of \MachineLearning models, demonstrating its potential for addressing data scarcity in \ChemicalEngineering and thus advance \MachineLearning on an industrial scale.

\section{Concept of Federated Learning}\label{text:sota}

\noindent In this section, we provide an overview of the concept of \FederatedLearning.
We focus on the scope of \ChemicalEngineering, where collaborating organizations, likely chemical companies that hold large data sets, jointly train a ML model while preserving data privacy, as illustrated in Figure~\ref{fig:FL_concept} and referred to as \emph{cross-silo} \FederatedLearning~\cite{Kairouz.2019, huang2022cross}. 
We further assume a \emph{centralized} collaboration, i.e., a central server/framework is used for communication between companies during model training, which is common for cross-silo \FederatedLearning~\cite{Kairouz.2019}.
For a thorough review of \FederatedLearning categorizations, we refer the interested reader to~\cite{Kairouz.2019,Liu.2024a,Wen.2023}. 

\subsection{Fundamentals \& Workflow}\label{text:sota_problem_formulation}

\noindent The term \FederatedLearning was introduced by McMahan et al.~\cite{McMahan.2017} and refined by Kairouz et al.~\cite{Kairouz.2019}: Multiple entities, referred to as clients, jointly solve a \MachineLearning task by model training without sharing their individual data.
For joint training, the clients do not exchange data to respect data privacy concerns.
They rather exchange information about the model and corresponding updates during training.
Therefore, \FederatedLearning shifts the focus from data sharing to model sharing.

\begin{figure}[hbt]
	\begin{center}
		\includegraphics[trim={1cm 1.5cm 1cm 3cm},clip, width=\textwidth, keepaspectratio]{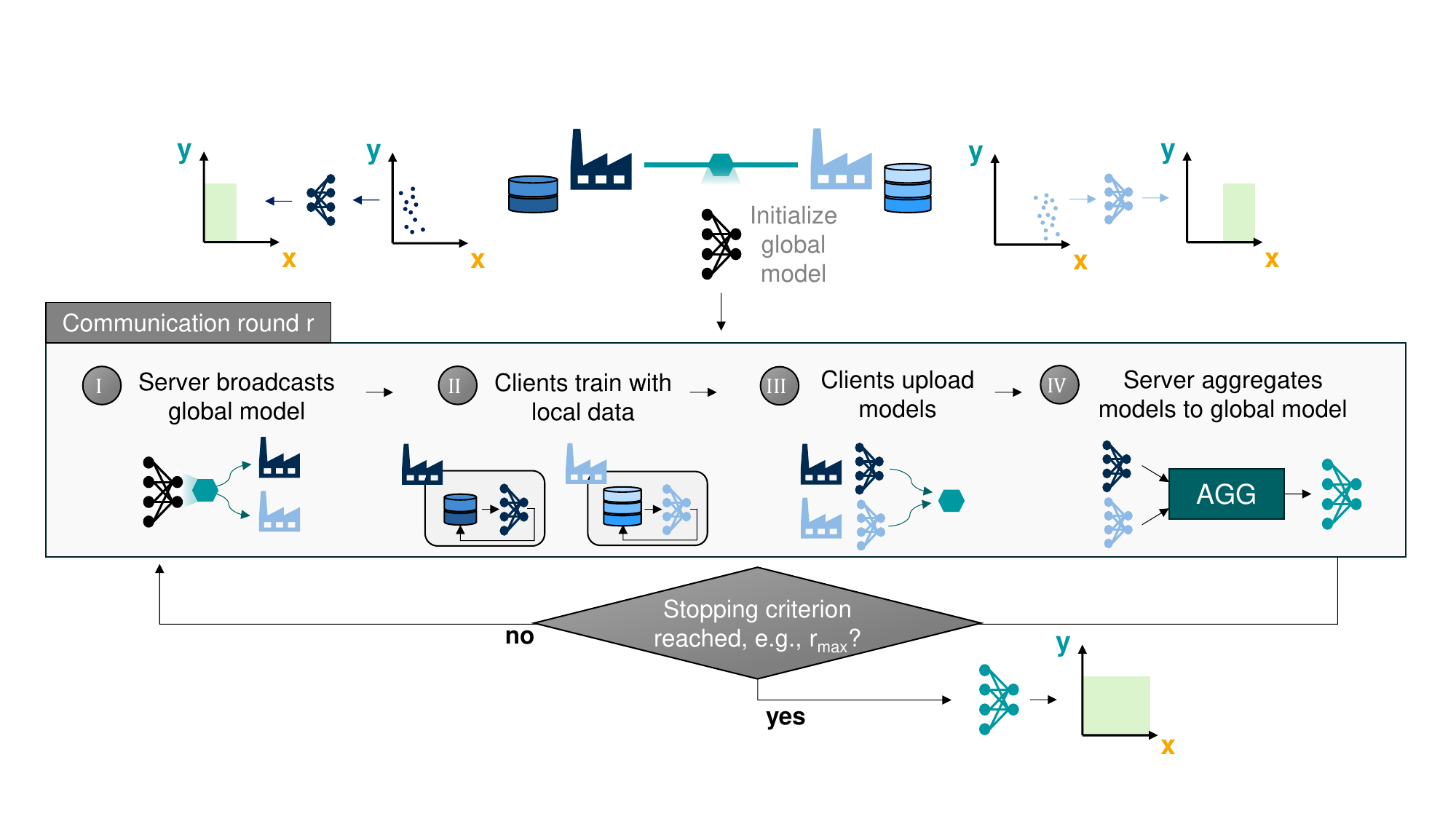}
		\caption{Illustrative workflow of federated learning for two companies jointly training a machine learning model.}
		\label{fig:FL_workflow}
	\end{center}
\end{figure}

We illustrate the workflow of \FederatedLearning in Figure~\ref{fig:FL_workflow}.
Based on~\cite{Dutta.2024, McMahan.2017, Kairouz.2019, Zhou.2024}, the setting can be formally described as follows:
$K$ clients aim to jointly develop a \MachineLearning model with model parameters $\mathbf{w} \in \mathbb{R}^d$.
Each client $k \in K$ holds a local data set $\mathcal{D}_k \sim \mathcal{P}_k(\mathbf{x},\mathbf{y})$ with a data point containing features $\mathbf{x}$ and corresponding labels $\mathbf{y}$, and $\mathcal{P}_k$ being the client's local data distribution. 
Instead of each client training an individual model as would be the case without \FederatedLearning and is indicated in the top left and right corners of Figure~\ref{fig:FL_workflow}, a global model is trained jointly.
Assuming that each client data distribution $\mathcal{P}_k$ is sampled independently and identically distributed (\textit{iid}) from a global data distribution $\mathcal{P}$, the joint training of the \MachineLearning model can be formulated as the following optimization problem~\cite{McMahan.2017, Zhou.2024}:
\begin{equation}\label{eqn:fl_general}
	\min_{\mathbf{w}} \ell(\mathbf{w})
	= \sum_{k=1}^{K} \frac{n_k}{n} \mathcal{L}_k(\mathbf{w}) 
	= \sum_{k=1}^{K} \frac{n_k}{n} \sum_{i=1}^{n_k} \ell_i(\mathbf{w}; (\mathbf{x}_i, \mathbf{y}_i) \in \mathcal{D}_k),
\end{equation}
where $\ell(\mathbf{w}) = \sum_{i} \ell_i(\mathbf{w}, (\mathbf{x}_i, \mathbf{y}_i))$ is the global loss function for training, and $\mathcal{L}_k(\mathbf{w})$ is the expected local loss function of client $k$ with $\mathbb{E}_{\mathcal{D}_k} [\mathcal{L}_k(\mathbf{w})] = \ell(\mathbf{w})$, in case of \textit{iid} local data sets. 
Further, $n_k = \# \mathcal{D}_k$ denotes the number of samples in the local data set of client $k$, so the total number of samples is $n = \sum_{k=1}^{K} n_k$.

To solve the optimization problem in Equation~\ref{eqn:fl_general}, an algorithm called Federated Averaging (FedAvg)~\cite{McMahan.2017} is commonly used.
In FedAvg, the training process starts by initializing the global model with parameters $\mathbf{w}_0$.
Then, the so-called communication rounds $r \in \{1, 2, ..., R\}$ are conducted.
In each communication round $r$~\cite{Kairouz.2019, Zhou.2024},
\begin{itemize}
	\item[(I)] the parameters of the global model from the previous communication round,  $\mathbf{w}_{r-1}$, are sent to the clients;
	\item[(II)] the clients update the received model parameters $\mathbf{w}_{r-1}$ by training with their local data set $\mathcal{D}_k$, i.e.,  $\mathbf{w}^k_{r} \in \argmin_{\mathbf{w}} \sum_{i=1}^{n_k} \mathcal{L}_k(\mathbf{w})$;
	\item[(III)] the clients upload the updated model parameters $\{\mathbf{w}^k_{r}\}_{k=1}^K$;
	\item[(IV)] then the updated model parameters of all clients are aggregated by applying an aggregation function to obtain new parameters of the global model $\mathbf{w}_{r} = \text{AGG}(\mathbf{w}^1_{r}, \mathbf{w}^2_{r}, ..., \mathbf{w}^{|K|}_{r})$, e.g., using weighted averaging: $\mathbf{w}_{r}~=~\sum_{k=1}^K \frac{n_k}{n} \mathbf{w}_{r}^k$, hence FedAvg~\cite{McMahan.2017}.
\end{itemize}
The new global model parameters are then passed to the clients as a starting point for a new communication round.
This communication process is repeated until a stopping criterion is met, e.g., a predefined maximum number of communication rounds, $R = r_{\text{max}}$, or convergence of the global model measured by changes in its parameters.

Joint training through \FederatedLearning promises higher model accuracies and applicability ranges than individual models.
The idea behind multiple rounds of communication is that through iterative sharing and retraining, the model parameters will converge to values that provide a good fit for all of the clients' data sets, i.e., the updated models by the clients (step II) reach a consensus~\cite{Kairouz.2019}.
In contrast to each client having an individual model with parameters fitted to its individual data set only, the models trained with \FederatedLearning are therefore influenced by more data and can result in higher accuracy and a wider applicability range (cf. green area in x-y plots in Figure~\ref{fig:FL_workflow}). 

We note that, next to FedAvg, a variety of other algorithms have been proposed for \FederatedLearning over the last years.
These algorithms typically share the goal of finding a parameter set that results in models with increased accuracy and larger applicability ranges for all clients. 
They can, for example, differ in the way the local models are aggregated and what kind of information is shared between the clients, e.g., sharing updates on model gradients instead of parameter values, which can result in improved convergence time and accuracy for different client and data distributions.
We herein focus on FedAvg, which has been successfully applied in various \FederatedLearning settings.
For an overview of different algorithms and convergence analysis, we refer the interested reader to~\cite{Kairouz.2019, Zhou.2024}. 

\subsection{Model Performance Indicators}\label{text:mpi}

\noindent To evaluate the model performance, e.g., in terms of accuracy and applicability range, the clients typically separate a test data set from their local data set $\mathcal{D}_k$, which is not used for training. 
So, standard metrics such as mean absolute error (MAE), mean squared error (MSE), and coefficient of determination ($R^2$) can be calculated for the global model that is passed to the individual clients.
Also, after the joint training, the clients can fine-tune the global model on their local data (similar to step II).
They can then compare the metrics for the global or fine-tuned model to a model trained only on their individual data.
For this, Heyndrickx et al.~\cite{Heyndrickx.2024} proposed the relative improvement of proximity to perfection (RIPtoP), which measures how much \FederatedLearning closes the gap between the baseline performance and a perfect model:
\begin{equation}\label{eqn:riptop}
	\text{RIPtoP} = \frac{\text{metric}_\text{MoI} - \text{metric}_\text{baseline}}{\text{metric}_\text{perfect} - \text{metric}_\text{baseline}},
\end{equation}

where the metric could for example be the MAE, $\textit{\text{MoI}}$ is the model of interest, which is the model after federated learning (the global or fine-tuned model), \textit{baseline} is the model trained only on client data set $\mathcal{D}_k$, and \textit{perfect} stands for an artificial perfect model. For example, $\text{metric}_\text{perfect}$ would be 0 if the metric is MAE and 1 if the metric is $R^2$. 

In research studies that explore \FederatedLearning in a simulated setting, the model \FederatedLearning performance is often compared to a \emph{centralized} training.
Centralized refers to training a model on the global data set $\mathcal{D}$, i.e., the aggregated set of the clients' local data sets.
This is indeed not possible in practical scenarios, where data sets cannot be aggregated due to privacy concerns.
Therefore, this comparison should be treated with caution as it simulates a scenario that is not practicable.
Also, the model performance in a centralized setting should not be seen as an upper bound, as it has been shown that the model trained in a federated setting can outperform the centralized setting, even if the overall data set remains the same~\cite{Dutta.2024, Zhou.2024}.

Another way to measure if clients benefit from \FederatedLearning can be to analyze the applicability domain of models~\cite{Heyndrickx.2023}. 
Further, it can be informative to perform sensitivity analyses with regard to several parameters, e.g., the number of clients, the level of data heterogeneity, and data quantity per client.

\subsection{Data Distribution \& Heterogeneity}\label{text:sota_heterogeneity}

\noindent The distributions of data sets across multiple companies typically differ in practice.
The problem formulation in Equation~\ref{eqn:fl_general} assumes that all client data distributions $\mathcal{P}_k$ are \textit{iid} samples from a global data distribution $\mathcal{P}$. 
However, several practical aspects typically lead to heterogeneous, non-\text{iid} data distributions, i.e., $\mathcal{P}_k \neq \mathcal{P}_l$ with $k \neq l$, also see~\cite{Kairouz.2019, Zhou.2024}.

\begin{figure}[hbt]
	\centering
	\begin{subfigure}[t]{0.3\textwidth}
		\centering
		\includegraphics[trim={13cm 6cm 13cm 6.5cm},clip, width=\textwidth, keepaspectratio]{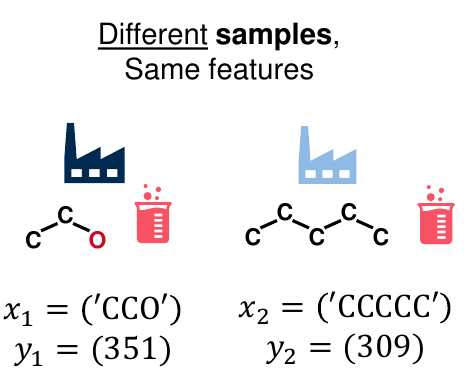}
		\caption{Horizontal}
	\end{subfigure}%
	\hfill
	\begin{subfigure}[t]{0.3\textwidth}
		\centering
		\includegraphics[trim={13cm 6cm 13cm 6.5cm},clip, width=\textwidth, keepaspectratio]{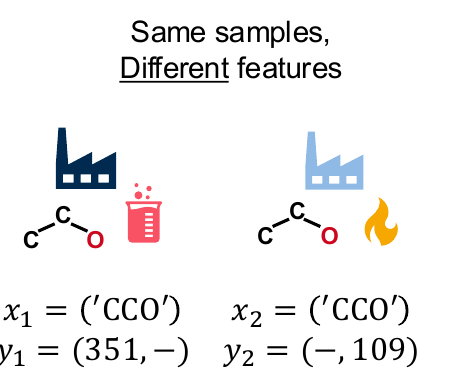}
		\caption{Vertical}
	\end{subfigure}%
	\hfill
	\begin{subfigure}[t]{0.3\textwidth}
		\centering
		\includegraphics[trim={13cm 6cm 12.5cm 6.5cm},clip, width=\textwidth, keepaspectratio]{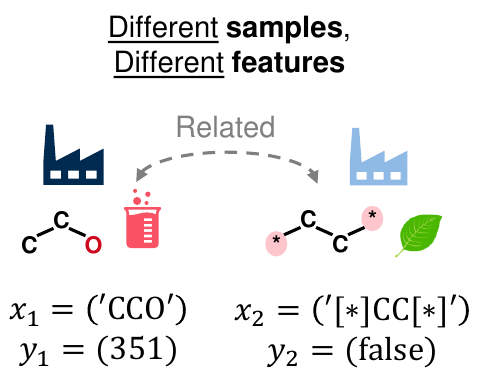}
		\caption{Transfer learning}
	\end{subfigure}%
	\caption{Data distributions for two companies at the example of molecular property data.}
	\label{fig:hetero_data}
\end{figure}

In \FederatedLearning, three types of data distributions are frequently considered, as shown with the example of molecular property data in Figure~\ref{fig:hetero_data}.
In \emph{horizontal} distributions, different companies have data sets with the same features, e.g., pure component boiling points with molecules represented as SMILES, but different samples, i.e., different pure components.
Vice versa, in \emph{vertical distributions}, the companies have data sets with the same samples, the same pure components, but different features, e.g., different properties such as boiling points and autoignition temperatures.
Lastly, the domains of the data sets can be related and enable \emph{transfer learning}; typically, both samples and features are different, e.g., one company has boiling point data for small molecules represented as SMILES, whereas another company has biodegradability data on polymers represented as BigSMILES.
Analogous examples can be derived for the process scale, e.g., (horizontal) two companies using a similar separation process for a specific mixture at different feed compositions and different operating conditions, (vertical) two chemical companies store material and process information along a product chain with one company being the raw material supplier and the other company being the manufacturer, and (transfer learning) two companies employing different types of reactors for chemically related reactions.

The different data distributions result in heterogeneous, non-\textit{iid} data.
Next to different features and samples, the data sets of the clients can differ in the mean and variance of the feature and label distributions, also resulting in non-\textit{iid} data.
In fact, practical scenarios often face combinations of the heterogeneity sources listed above. 

From an optimization perspective, the non-\textit{iid} data sets can result in the local loss functions deviating from the global loss function, such that $\mathbb{E}_{\mathcal{D}_k}~[\mathcal{L}_k(\mathbf{w})]~\neq~\ell(\mathbf{w})$~\cite{Yang.2024}. 
This can limit parameter convergence in training and thus have a negative impact on model performance.
To address data heterogeneity, pre-training~\cite{Nguyen.2023b}, multi-task, and transfer learning~\cite{Smith.2017, Qi.2024} can be used.
Also, so-called \emph{personalized} \FederatedLearning~\cite{Dinh.2024}, which balances the performance of local and global models, can be employed.
Empirically, common \FederatedLearning algorithms such as FedAvg have often shown robust performance -- even in heterogeneous data settings, cf.~\cite{Kairouz.2019, Zhou.2024, shao2023survey, wang2022unreasonable}.

\section{Towards Federated Learning in Chemical Engineering}\label{text:FLCE}

\noindent We now move to \FederatedLearning for \ChemicalEngineering. 
Having discussed existing applications in Section~\ref{sec:Intro}, we here focus on perspective \FederatedLearning applications (Section~\ref{text:FLCE_applications}) and corresponding challenges (Section~\ref{text:FLCE_challenges}) within the chemical industry. 

\subsection{Perspective Applications}\label{text:FLCE_applications}

\begin{figure}[tphb]
	\begin{center}
		\includegraphics[trim={0cm 4cm 0.5cm 2cm},clip, width=0.9\textwidth, keepaspectratio]{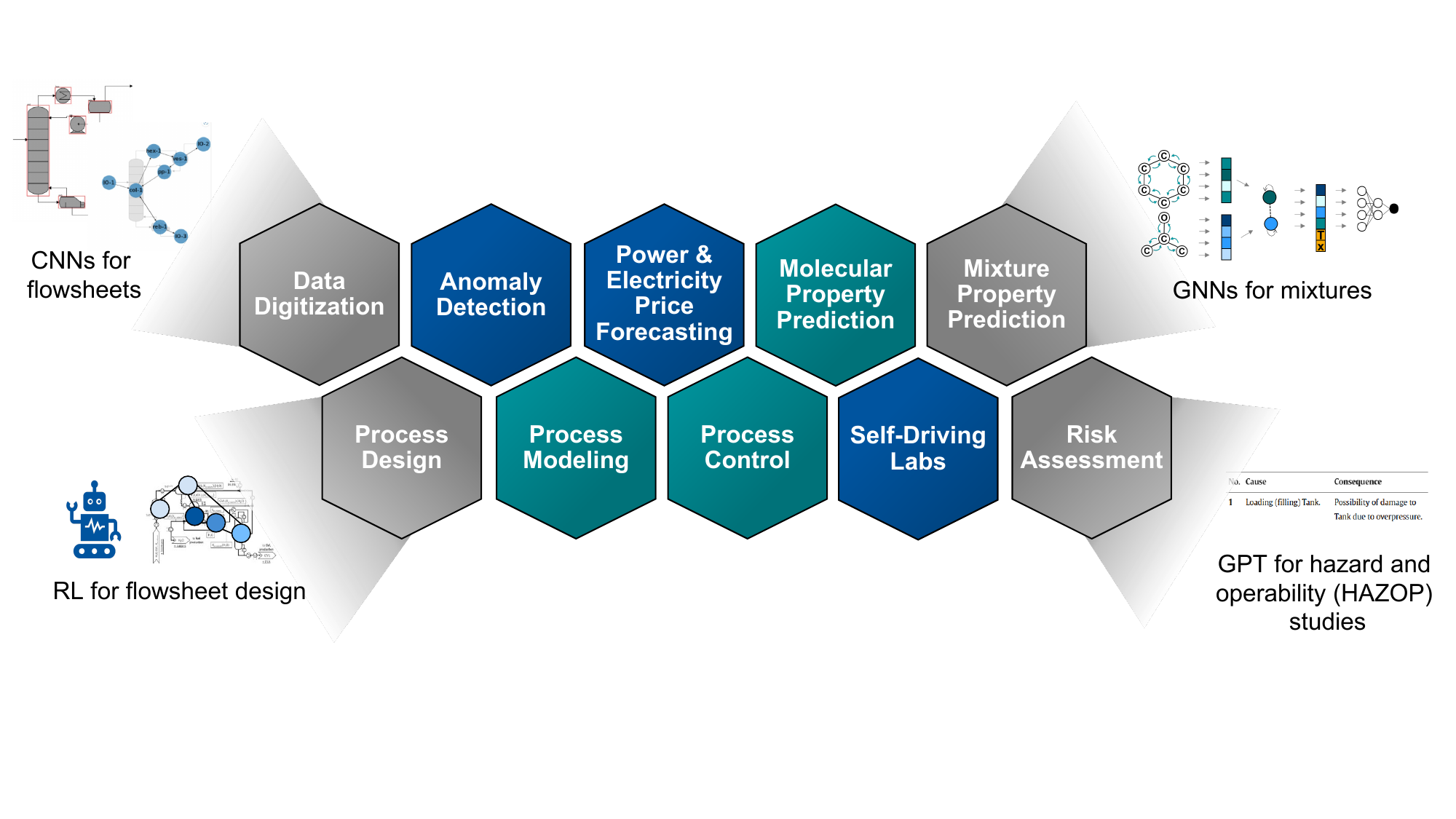}
		\caption{Potential application fields of federated learning in chemical engineering. The colors indicate the current status in the literature: (green) has already been applied, (blue) has been proposed, or (grey) is rather unexplored. We further show potential applications within the unexplored fields; images adapted from~\cite{Rittig.2024, theisen2023digitization, stops2023flowsheet, Ekramipooya.2024}.}
		\label{fig:potential_applications}
	\end{center}
\end{figure}

\noindent We envision that \FederatedLearning enables chemical companies to collaborate in the development of \MachineLearning models while respecting data privacy.
We show perspective \FederatedLearning applications within the chemical industry in Figure~\ref{fig:potential_applications}.
Note that we do not aim at a comprehensive review of \MachineLearning applications in \ChemicalEngineering or chemistry as such overviews can be found elsewhere~\cite{Schweidtmann.2021, StriethKalthoff.2020, Daoutidis.2024, Cheng.2024}. 
Rather, we highlight applications for which we think that \FederatedLearning can make a substantial impact based on readily available \MachineLearning methods.
We envision that this list will be extended in the future with further developments in \MachineLearning for \ChemicalEngineering.

\textbf{Data Digitization:} Digitizing historical and newly obtained data is of major interest in the chemical industry.
Many chemical enterprises hold large amounts of chemical and process data and are currently in the process of digitizing this data and corresponding knowledge. 
This includes the digitization of process descriptions and flowsheets, historical process data, and setups and results from lab experiments. 
To support this process, \MachineLearning tools based on computer vision can be utilized, for example, for digitization of process flowsheets~\cite{theisen2023digitization, kang2019digitization, Sturmer.2024, Balhorn.2022}. 
However, openly accessible flowsheets are mostly restricted to literature sources~\cite{Balhorn.2022}. 
Companies are typically not willing to share their flowsheets publicly. 
Therefore, \FederatedLearning could be utilized to enable joint training of such models.
In general, using larger and more diverse data from the chemical process domain through \FederatedLearning will facilitate training ML models for digitizing chemical data without the need to share proprietary data.

\textbf{Process Modeling:} Going one step further, ML is greatly impacting the modeling of chemical processes in recent years~\cite{Schweidtmann.2021, Daoutidis.2024}.
Specifically, ML is often used in surrogate modeling~\cite{bhosekar2018advances, schweidtmann2019deterministic, mcbride2019overview, esche2022architectures, misener2023formulating} and hybrid modeling~\cite{bradley2022perspectives, schweidtmann2024review, sansana2021recent} to capture the behavior of chemical phenomena, unit operations, or entire processes that are not yet fully understood mechanistically and/or have a high mathematical complexity.
Furthermore, ML can guide the modeling of chemical processes.
For example, ML models have been used to determine reaction networks~\cite{meuwly2021machine, margraf2023exploring, weber2021chemical}, rates~\cite{jorner2021machine, chung2024machine}, and kinetic models~\cite{quaglio2020artificial}.
However, the data available for process modeling is typically limited in terms of amount but also in terms of variation and diversity.
Therefore, ML applications in process modeling typically face scarce data for similar conditions of a single phenomenon, unit operation, or entire process, and thus do not allow generalization beyond, which also hinders subsequent process optimization.
The reasons for these data limitations can be manifold, including a low level of digitization, cost-intensive measurements and experiments, and narrow operation ranges of processes in practice, e.g., due to safety constraints.
Here, \FederatedLearning has high potential to make use of data on similar applications from multiple companies, thereby allowing ML models to learn from larger and more diverse data sets.
As highlighted in the introduction, Hayer has indicated this potential by using \FederatedLearning to train a neural network on aluminum electrolysis data from different chemical producers~\cite{Hayer.2021}.
Another example for future work would be a reaction kinetic classifier that is trained jointly by multiple companies on their proprietary kinetic data of different reactions.
This will lead to enhanced robustness and generalization capabilities of ML-based chemical process models.

\textbf{Process Design:} Following the modeling stage, ML approaches have recently been transferred to process design~\cite{schweidtmann2024generative}.
Specifically, generative ML has been utilized to design~\cite{gottl2022automated, Vogel.2023, mann2024esfiles} and automatically correct~\cite{Schulze-Balhorn.2024} process flowsheets and predicting their control structures~\cite{balhorn2025graph}. 
For this, mainly transformer and reinforcement learning models have been proposed, cf.~\cite{Vogel.2023, stops2023flowsheet, gao2024deep, gottl2025deep}.
Such models can highly benefit from (pre-)training on large datasets.
However, as mentioned above, public process data, e.g., in the form of flowsheets, is highly limited.
This also results in the limitation of current models being typically trained for a specific process design case study~\cite{gao2024deep}.
That is, the generative models hardly generalize and need to be retrained for new design tasks.
Generalization in ML-based process design is thus an active research field, see, e.g.,~\cite{gottl2025deep}.
Using \FederatedLearning for joint training of generative models for flowsheet design by several chemical companies enables learning from a more diverse set of processes, promising to advance generalization capabilities.

\textbf{Process Control:} Using ML for control of chemical processes has been an active research field in recent years~\cite{Daoutidis.2024, tang2022data}.
ML applications in control include system identification by capturing non-linear dynamics through neural networks~\cite{esche2022architectures, chiuso2019system, ahmed2025comparative}, and estimation of optimal control policies by reinforcement learning~\cite{hoskins1992process, badgwell2018reinforcement, nian2020review, pr13061791}.
As dynamic system responses and corresponding optimal control highly depend on the specifics of the process at hand, jointly training ML models based on different processes is challenging.
Yet, in recent years, transfer learning has gained increasing attention, that is, to transfer knowledge from (multiple) source processes to new, related processes, see, e.g.,~\cite{xiao2023modeling, ArceMunoz.2024, Xiao.2024}. 
Such knowledge transfer promises to reduce modeling and system identification efforts for control, e.g., by reducing required data amounts.
While companies presumably cannot share dynamic process data and control strategies due to privacy concerns, \FederatedLearning provides an opportunity to jointly train ML models for transfer learning purposes.

Furthermore, Xu~\&~Wu~\cite{Xu.2024} recently demonstrated the use of \FederatedLearning for distributed model predictive control of a reaction network, in which information exchange between reactors is limited.
In general, control in network structures, where multiple entities with proprietary process information participate and could benefit from cooperation, such as in distributed energy resource management systems for utilities, power and microgrids~\cite{morstyn2016control, otashu2021cooperative, dobbe2019toward}, bear large potential for \FederatedLearning, with applications currently emerging~\cite{lee2020federated, li2023federated, joshi2023survey}. 
An interesting aspect here is to consider the close relationship of \FederatedLearning with algorithms for distributed optimization in process networks~\cite{Allman.2022, Li.2025}. 

\textbf{Anomaly Detection:} Data on anomalies in real processes is naturally scarce, limiting \MachineLearning applications. 
\MachineLearning approaches have been actively explored to detect anomalies and faults in chemical processes and beyond~\cite{qin2019advances, choi2021deep, Hartung.2023}.
These approaches cover classification methods, such as support vector machines (SVMs), to distinguish normal process conditions from anomalies, and sequence models, such as recurrent neural networks, to analyze historical process time-series data, cf. overviews in~\cite{Hartung.2023, arunthavanathan2021analysis}.
In particular, the Tennessee Eastman process~\cite{Downs.1993} serves as a benchmark for anomaly detection on process time-series data, which has been tackled by various \MachineLearning models including generative adversarial networks~\cite{yang2019generative}, recurrent neural networks~\cite{lomov2021fault}, GNNs~\cite{jia2023topology}, and transformers~\cite{schoch2024deep}, also cf.~\cite{Hartung.2023} for a comprehensive comparison.
However, the corresponding data set is synthetic and limited to the Tennessee Eastman process.
The detection of anomalies in real process data is presumably more challenging~\cite{Hartung.2023}.
Further, it is desirable to apply anomaly detection methods across a variety of different processes.
It has recently been shown that transfer learning, using data from related tasks, can be beneficial in \MachineLearning applications for anomaly detection in industrial time-series data~\cite{Yan.2024}.
Having more real-world data from related processes usable for training through \FederatedLearning could lead to deep anomaly detection algorithms that have higher detection ratios and generalize across multiple process types.
Thus, we envision that \FederatedLearning can enhance predictive maintenance and process safety. 

\textbf{Risk Assessments:} The creation of risk assessments of chemical processes and plants, especially for hazard and operability (HAZOP) studies, is a critical but time-consuming task. 
Over recent years, data-driven approaches have been explored to support risk assessment studies.
This includes the aforementioned ML models to predict failures and anomalies from process time-series data.
Furthermore, ML approaches from natural language processing, such as large language models (LLMs), have been used to extract information from chemical accident databases~\cite{single2020knowledge} and facilitate HAZOPs~\cite{feng2021application, Ekramipooya.2024}. 
A particular challenge is that relevant data for HAZOPs typically consists of different modalities, mainly text, images, and tables~\cite{Niu.2024}. 
Recent advances in ML provide methods that can utilize such multiple modalities, cf.~\cite{baltruvsaitis2018multimodal, liang2024foundations, wu2024next}.
The bottleneck for training associated models on chemical processes is the insufficient amount of publicly available data, as HAZOP studies from chemical enterprises are protected due to intellectual property and safety-critical information~\cite{Niu.2024}. 
We envision the training of multimodal ML models for risk assessments that can deal with tabular, text, and image data. 
As such multimodal models need vast amounts of training data -- even when the training is started from pre-trained LLMs--, it is desirable to leverage the process safety data from several chemical companies for training.
\FederatedLearning presents a way to train such models, making use of industrial risk assessments without exposure of the underlying data.

\textbf{Power and Electricity Price Forecasting:} The future energy system will be decentralized and based on renewable, fluctuating energy sources~\cite{Scholtz.2024}. 
This gives rise to the field of forecasting time-series data like energy prices and power generation, which are highly relevant for the chemical industry.
Due to the decentralized nature of future energy systems, \FederatedLearning can be a key technology to combine energy data from different sources while keeping such data private~\cite{Zhang.2025}.
In a recent review, ElRobrini et al. discuss such \FederatedLearning-based approaches in the light of photovoltaic and wind power applications.
We envision that enhancements in power and electricity price forecasting through \FederatedLearning will be beneficial for data-driven methods used in tackling flexibilisation of the chemical industry, like planning and scheduling, demand-side management, and economic nonlinear model predictive control~\cite{Tsay.2019}. 

\textbf{Molecular Property Prediction:} Moving from the process and systems scale to the molecular scale, we see large potential of \FederatedLearning to develop ML models for molecular property prediction, such as graph neural networks~\cite{Coley.2017, Rittig.2022, Reiser.2022, Felton_MLSAFT.2023, Heid.2024} and transformers~\cite{chithrananda2020chemberta, chen2023generalizing, winter2025understanding}.
In fact, the application of \FederatedLearning for learning tasks related to molecular data is increasingly targeted in pharmaceutical research~\cite{Zhu.2022, Chen.2021, Huang.2023}.
Specifically, the idea is that multiple companies join a collaboration in training ML models to predict drug properties, whereas the proprietary molecular property data that has been gathered over the last decades by each company stays private.
This allows for the enlargement of both the coverage of the chemical space and the property range used for training beyond publicly available datasets or individual datasets owned by the companies.
In doing so, the aforementioned MELLODDY project has demonstrated a successful application of \FederatedLearning on an industrial scale with multiple pharmaceutical companies~\cite{Heyndrickx.2024}.
With the MELLODDY project serving as a blueprint, we advocate for similar collaborative efforts in \FederatedLearning projects for molecular property prediction by the chemical industry.
As an example, we envision that ML models will be able to predict a wide range of thermodynamic properties for a wide spectrum of molecular classes that are relevant for chemical process design and optimization.

In conjunction with the improved prediction of properties, \FederatedLearning has recently been studied for computer-aided molecular design~\cite{Manu.2024, Guo.2024} and multimodal molecular representation learning~\cite{Feng.2024}. 
These works can serve as starting points for the collaborative development of data-driven molecular design models~\cite{gomez2018automatic, Bilodeau2022, decardi2024generative}, which can, for example, be used for solvent~\cite{zhang2023deep, pirnay2025graphxform}, fuel~\cite{RittigRitzert_GraphMLFuel.2022}, and catalyst~\cite{dos2021navigating, ishikawa2022heterogeneous, schilter2023designing} design, in the chemical industry.

\textbf{Mixture Property Prediction:} Building on ML-based property prediction for pure components, great efforts have recently been dedicated to predicting properties of mixtures, which are highly relevant for chemical processes.
Here, ML models, such as graph neural networks~\cite{SanchezMedina.2022, Rittig_GNNgammaIL.2022, Vermeire.2021, Qin.2023, leenhouts2025pooling}, matrix completion methods~\cite{Jirasek.2020, chen2021neural}, and transformers~\cite{Winter.2022}, have been trained on quantum mechanical data, publicly available datasets, or commercial databases.
Beyond these data, chemical companies have created private property databases of molecules and their mixtures over the last decades.
Analogously to pure component properties, we envision that \FederatedLearning is used by academia, database organizations, and chemical companies to jointly develop ML models, without the need to share their property data.
This will allow to utilize additional combinations of molecules and corresponding mixture properties during training and thereby enhancing predictive capabilities.

\textbf{Self-Driving Labs:} The advent of self-driving labs promises to catalyze chemical discovery and reaction optimization~\cite{abolhasani2023rise, Tom.2024}. 
The parallel execution of lab tasks by multiple self-driving labs located in different regions of the world has been actively studied~\cite{Skilton.2015, Fitzpatrick.2018, Bai.2024}.
As data-driven models become an integral part of such self-driving labs, e.g., for automated synthesis prediction, spectra analysis, and experimental planning, we see that \FederatedLearning provides a way to use larger datasets for model training, especially if standardized protocols for self-driving labs mature, also cf.~\cite{Tom.2024}. \newline

\noindent Overall, we find great potential of \FederatedLearning for numerous applications in \ChemicalEngineering, ranging from the molecular to the process scale.
We advocate for collaborative efforts between the chemical industry, database organizations, and academia to adapt and apply \FederatedLearning, thereby advancing \MachineLearning models without the need for conducting expensive experiments to obtain new data -- all while respecting data privacy.
Ultimately, we envision using \FederatedLearning to leverage the substantial data and knowledge repositories of chemical companies to develop and fine-tune multi-purpose \MachineLearning models \cite{Li.2024, Bommasani.2021, Das.2024, DecardiNelson.2024, Ren.2024, bran2023chemcrow, choi2025perspective}, also known as foundation models, for the field of \ChemicalEngineering, cf.~\cite{Sani.2024a}.

\subsection{Challenges \& Research Directions}\label{text:FLCE_challenges}

\noindent Initiating and implementing \FederatedLearning in projects within the chemical industry comes with several challenges.
Next, we discuss these challenges and provide related research directions.

\textbf{Incentives:} The basic premise for \FederatedLearning and corresponding potential improvement of \MachineLearning models is that chemical companies and further data holders enter into collaborations. 
However, these companies are competitors -- typically with conflicting interests --, so in most cases, they are not willing to openly share their proprietary data with others.
Consequently, the initiation of \FederatedLearning projects inherently poses a trade-off between cooperation and competition~\cite{Huang.2022b} and is thus challenging.
A common approach to analyze this trade-off is game theory, which has already been applied to deduce incentives for \FederatedLearning projects in general~\cite{Tu.2022, Huang.2022a, Huang.2024c}.
Here, a major aspect is to value the data that clients contribute to increase the model performance~\cite{Wu.2024} because the core principle of \FederatedLearning is that clients' data cannot be accessed by other clients. 
For example, one client might contribute more data than another client and will therefore expect a higher reward than the other client for taking part in the training process~\cite{Sim.2024}. 
Thus, ensuring a fair \FederatedLearning process with corresponding incentives is challenging~\cite{Huang.2024d}.
Game theoretical approaches in \FederatedLearning are so far restricted to a few competitors~\cite{Huang.2024a} and have not been investigated for specific characteristics of the chemical industry, therefore requiring further research.
Most importantly, demonstrating the benefits of \FederatedLearning for exemplary \ChemicalEngineering tasks, also in simulated settings, is needed to incentivize chemical companies to initiate \FederatedLearning projects.

\textbf{Adversarial Attacks:} Potential data privacy threats can discourage clients from participating in \FederatedLearning projects.
Adversaries might attempt to reverse engineer data samples based on the shared global model parameters and corresponding model updates.
For example, Kr{\"u}ger et al. recently investigated the possibility of extracting information on molecular structures present in the data sets of individual clients from a model trained through \FederatedLearning for property prediction in drug applications~\cite{kruger2025publishing}.
Further, clients could try to introduce bias into the model to deteriorate the predictive performance of other clients, referred to as data and model poisoning~\cite{bagdasaryan2020backdoor}.
Therefore, data privacy and security are an active and crucial field of research in \FederatedLearning~\cite{Zhao.2024}.
While recent works have investigated adversarial attacks on ML models in \ChemicalEngineering by external parties, e.g., for process control~\cite{koay2023machine} and demand-side management~\cite{Cramer.2024}, such investigations are lacking for \FederatedLearning setups, particularly at the process scale, and require further research.

\textbf{Data Heterogeneity \& Uncertainty:} The majority of chemical and process data will naturally show some level of heterogeneity and uncertainty. 
This originates from chemical companies working on different regions of the chemical space, operating processes with different raw materials and at different conditions in varying demographic locations, and using different types and quality of equipment, e.g., for sensors. % with different measurement uncertainties or noise.
Results from lab experiments will differ by experimental uncertainty, and process data will show measurement noise, referred to as aleatoric noise, cf.~\cite{Heid.2023}.
In such cases, data samples might share the same features but have different labels, also cf. Section~\ref{text:sota_heterogeneity}. 
Hence, dealing with data heterogeneity is an inherent challenge in \FederatedLearning research for chemistry and \ChemicalEngineering.
We anticipate that data availability from various sources will enable the refinement of aleatoric uncertainties in chemical and process data and of corresponding uncertainty quantification methods in a federated setting, similar to~\cite{linsner2021approaches, zhang2023uncertainty}.
Furthermore, research on \FederatedLearning has recently addressed the field of heterogeneity and label noise across clients, e.g., in~\cite{tuor2021overcoming, Fang_2022_CVPR, yang2022robust, tsouvalas2024labeling}, which can be a starting point for further investigations in \ChemicalEngineering.

Furthermore, we emphasize that data formats can differ greatly across chemical companies. 
This poses the challenge of deciding on input formats of the data used for training an ML model with \FederatedLearning.
However, since data is not collected and stored centrally, but rather remains within individual companies, they can continue working with their existing data formats and implement parsers within the \FederatedLearning framework.

\textbf{Benchmarks:} To drive research in \FederatedLearning for \ChemicalEngineering, we advocate to create open-source benchmarks.
This will require the collection and publication of data sets by experts from academia and industry in order to cover realistic data distribution scenarios across chemical companies.
Based on these benchmarks, leaderboards should be introduced to make different \FederatedLearning approaches comparable in a straightforward manner.
Developed \FederatedLearning models could also be evaluated by individual companies on their proprietary test data sets, with only the evaluation metrics shared publicly.
As the research for \FederatedLearning in \ChemicalEngineering is at an early stage, designing such benchmarks can facilitate establishing data characteristics and research needs relevant in the chemical industry. \newline

\noindent Overall, \FederatedLearning comes with an interplay of various aspects, such as data heterogeneity, privacy concerns, and incentives, and thus requires further research for \ChemicalEngineering applications in joint efforts of academia and industry.
Practical implementations of \FederatedLearning projects also require setting up computational infrastructures and coding frameworks under consideration of cyber security and legal aspects, cf.~\cite{Woisetschläger.2024a, Chik.2024}.
For this, the chemical industry can use successful large-scale industrial projects in related domains~\cite{Heyndrickx.2023, Oldenhof.2023, aibs.2024}.

\section{Exemplary Case Studies}

\begin{figure}[h!bt]
	\centering
	\includegraphics[width=\linewidth]{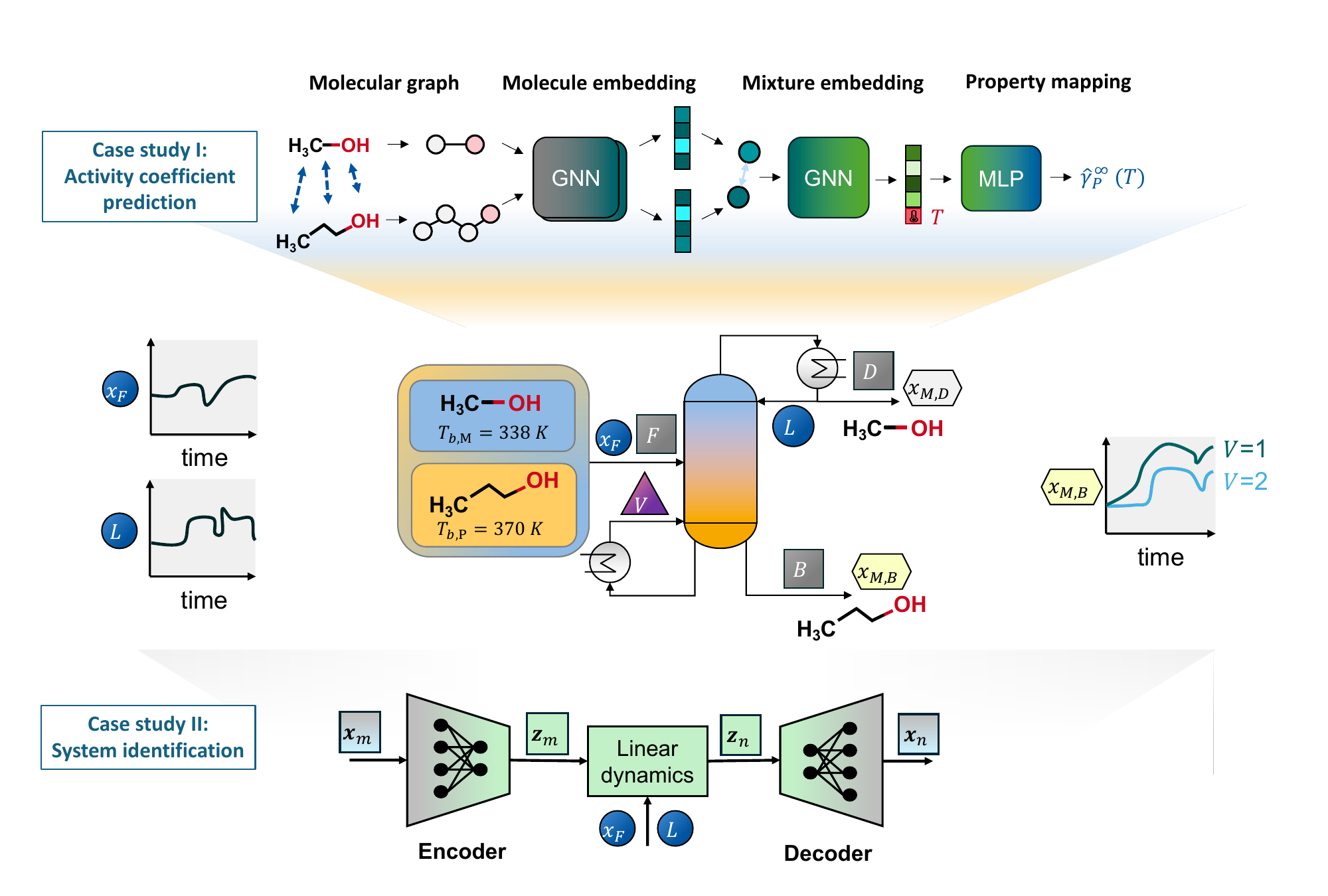}
	\caption{Illustration of process for separating methanol and propanol by means of a distillation column with two related chemical engineering tasks: (I) predicting activity coefficients of binary mixtures by graph neural networks (GNNs), and (II) identifying a low-dimensional model for predicting system responses for dynamic input flows.}
	\label{fig:case_studies_overview}
\end{figure}

\noindent In this section, we demonstrate the application of \FederatedLearning in \ChemicalEngineering.
As illustrated in Figure~\ref{fig:case_studies_overview}, we consider two case studies motivated by separation processes, covering the molecular scale and the process scale.
In the first case study, we investigate \FederatedLearning for the molecular scale by training graph neural networks that predict mixture properties (Section~\ref{text:gnn}).
In the second case study, we explore \FederatedLearning for the process scale by training a data-driven model that predicts dynamic system responses of a distillation column with varying feed composition (Section~\ref{text:mpc}).

\subsection{Case Study I: Federated Learning of Activity Coefficients in Binary Mixtures}\label{text:gnn}

\noindent We first investigate \FederatedLearning of GNNs for predicting activity coefficients in binary mixtures at infinite dilution, which are particularly relevant for chemical separation processes.
\FederatedLearning has already been applied to GNNs, cf.~\cite{Liu.2024b}, also for prediction of pure component properties, e.g., phyiscochemical properties~\cite{Zhu.2022} and drug properties~\cite{Kruger.2024}.
However, the application to mixture properties relevant to \ChemicalEngineering is missing so far.

\subsubsection{Model}
\noindent The prediction of activity coefficients in binary mixtures with \MachineLearning is an active area of research~\cite{Jirasek.2020, Winter.2022, Winter.2023, specht2024hanna, ottaiano2024machine}, with GNNs being a prominent method, cf.~\cite{Qin.2023, Felton.2022, SanchezMedina.2023, Rittig.2023a, Rittig.2023b, Rittig.2024}.
As shown in Figure~\ref{fig:case_studies_overview}, GNNs use graph convolutions to learn a fingerprint vector of a molecule based on its graph representations, which is then mapped to a property of interest by a multilayer perceptron~(MLP)~\cite{Rittig.2022, Reiser.2022, Gilmer.2017}.
For predicting activity coefficients of binary mixtures, we consider a GNN model from our previous work~\cite{Rittig.2023b}: 
To handle mixtures, we take the two molecular graphs of the solute and solvent as input and combine their fingerprints to a mixture fingerprint in an aggregation step, which is based on a mixture graph, cf.~\cite{Qin.2023, Rittig.2023b}.
The mixture fingerprint is then mapped to the activity coefficients. 

\subsubsection{Data Set \& Federated Learning Scenarios}\label{text:gnn:data}
\noindent We use the infinite dilution activity coefficient dataset published by Brouwer et al.~\cite{Brouwer.2019} in the sanitized version by Winter et al.~\cite{Winter.2022} and Sanchez Medina et al.~\cite{SanchezMedina.2023}. 
The data set consists of 18,016 activity coefficient samples of binary mixtures at infinite dilution at varying temperatures. 

We use this dataset as a starting point to create two federated learning scenarios in a setting with four companies:
\begin{itemize}
	\item \textbf{even-random (whole model shared)}: We randomly partition the data set into subsets of equal size (25\%) and assign each to a company.
	Thus, the companies' data sets tend to be distributed identically and independently (\textit{iid}), cf. Section~\ref{text:sota_problem_formulation}, allowing us to validate \FederatedLearning in a rather ideal setting.    
	Here, we assume that the companies share the whole model, and we use the FedAvg algorithm for training.
	\item \textbf{uneven-scaffold (partial model shared)}: We partition the data based on solvent scaffolds with uneven partition sizes of 40\%, 30\%, 20\%, and 10\% for the companies, respectively.
	Using scaffolds captures that different companies work on different regions of the chemical space. We partition the data based on scaffolds of the solvents. We find that acyclic molecules represent the most common scaffold, such that both clients 1 and 2 have partitions of solely acyclic solvents, whereas the partitions of clients 3 and 4 consist of different scaffolds.
	Inspired by the MELLODDY project \cite{Heyndrickx.2023}, we assume that the companies do not share the whole model, but only the GNN layers up to the mixture embedding in order to increase their privacy and decrease risks of adversarial attacks (cf. Section \ref{text:FLCE_challenges}).
	That is, the prediction MLP is not shared and remains private, referred to as FedPer, see~\cite{Arivazhagan.2019, Pillutla.2022}.
	This scenario covers several heterogeneity aspects, cf. Section~\ref{text:sota_heterogeneity}, which we expect in practical data and thus enables evaluating \FederatedLearning in a more realistic setting. 
\end{itemize}

As baselines, we consider (i) each company training a model only on their \textbf{private data} set and (ii) training a model on the \textbf{full data} set of activity coefficient, hence a centralized setting \ref{text:mpi}. 
In each case, we split the (partitioned) data sets into 70\% training, 15\% validation, and 15\% test set.

To assess the model quality, we report the following metrics. 
We compute the test set MSE after local training before \FederatedLearning aggregation, i.e., for the local models, and after \FederatedLearning aggregation, hence for the global model. 
Further, we compute the RIPtoP metric, cf. Equation~\ref{eqn:riptop}, where we use the test set MSE after the first round of local training as the \textit{baseline} metric and the test set MSE after the last round of aggregation as \textit{MoI} metric.

\subsubsection{Implementation \& Hyperparameter}
\noindent We use the GNN implementation of our MoLprop framework~\cite{Rittig.2022}, which is based on PyTorch Geometric~\cite{Fey.2019} and the RDKit~\cite{rdkit}.
We use the default hyperparameters from MoLprop throughout all experiments and only remove the chirality atom feature because it has no variation in the data set. 

For the implementation of the \FederatedLearning algorithms, we use Flower 1.7~\cite{Beutel.2020}.
Throughout all experiments, we use four clients, assume that all clients are connected at every communication round, and use the same hyperparameters for each client. 
We use 30 communication rounds and 150 local epochs with exponential learning rate decay and early stopping. 
We conduct each of our experiments ten times and report the arithmetic mean together with the standard error.

\subsubsection{Results \& Discussion}\label{text:gnn:results}
\begin{figure}[htpb]
	\centering
	\begin{subfigure}[b]{0.5\textwidth}
		\centering
		\includegraphics[width=\linewidth]{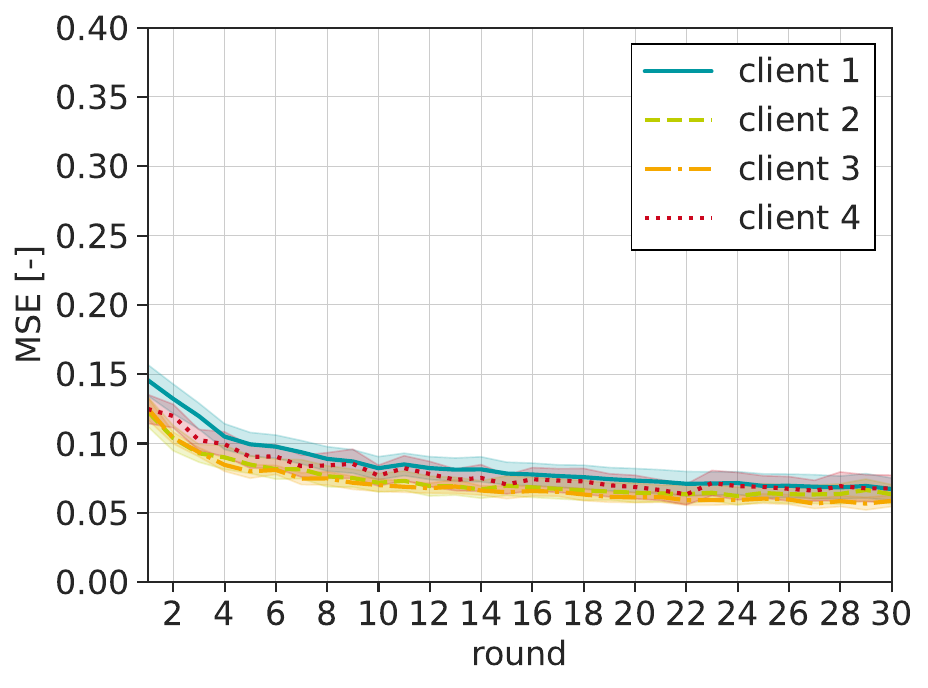}
		\caption{}
		\label{fig:random_clients}
	\end{subfigure}\hfill
	\begin{subfigure}[b]{0.5\textwidth}
		\centering
		\includegraphics[width=\linewidth]{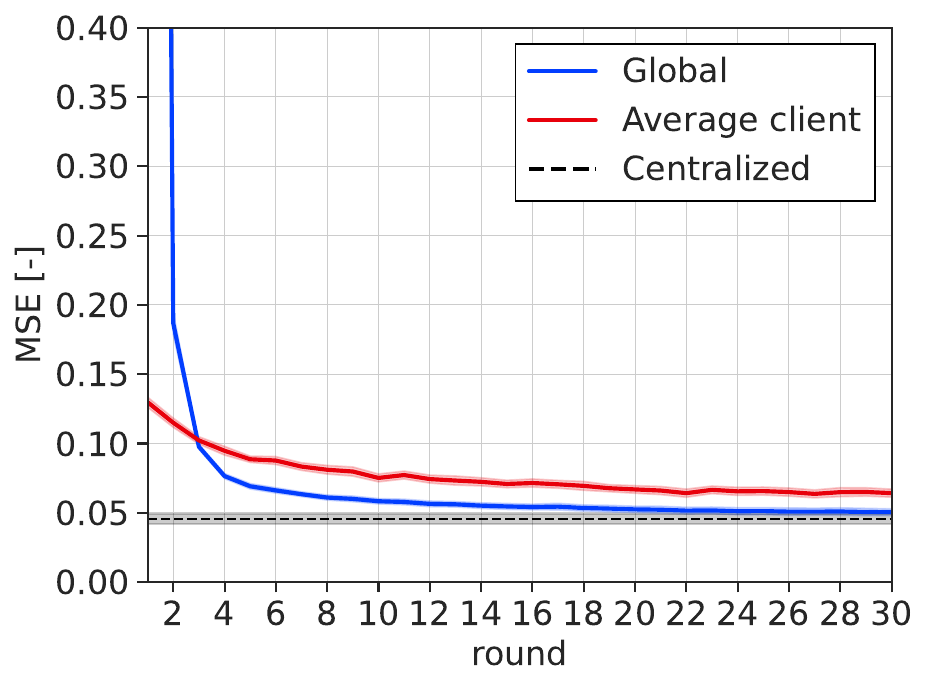}
		\caption{}
		\label{fig:random_all}
	\end{subfigure}
	\caption{The test mean squared error (MSE) values for the \textbf{even-random (whole model shared)} scenario: \textbf{(a)} results for the respective model of the \emph{individual clients}; \textbf{(b)} aggregated results for the \emph{global} model that is obtained after each round through aggregation of the individual clients' models, the \emph{average client}, i.e., the averaged performances of the individual clients' models from (a), and \emph{centralized} model that is trained on the whole data set data and serves as a baseline.}
	\label{fig:random}
\end{figure}

\noindent We first report and discuss the results for the even-random scenario, where the whole model is shared among the clients.
An important question is whether \FederatedLearning provides benefits for the clients in comparison to training just on their own data.
In Figure~\ref{fig:random_clients}, the test MSEs for the models of the four individual clients are shown over the communication rounds.
These models are evaluated after individual model training by the clients in each communication round, thus before model aggregation.
The MSE values in the first round range from 0.123 to 0.146 and represent the performance that the clients get without applying any \FederatedLearning (as the communicated global model in the first round is initialized randomly).
In the following communication rounds, the prediction errors decrease significantly for all clients.
Specifically, after 30 rounds, the MSE values range from 0.059 to 0.068; that is, the error range is more than halved.
Therefore, \FederatedLearning yields a consistent performance gain for the clients in the even-random scenario.

In Figure~\ref{fig:random_all}, we further show the average test MSE of the individual clients (in red) in comparison to the \emph{global} model (in blue), which corresponds to the aggregated \FederatedLearning model after each communication round.
Notably, the test MSE of the global model after the first communication is very high with a value of 2.770, which can be explained by the fact that the models trained by the individual clients are not calibrated to each other, so aggregating these highly decreases the prediction performance.
Through the following communication and retraining of the models, the test MSE of the global model decreases drastically and starts to converge, reaching a value of 0.050 after 30 rounds, yielding a RIPtoP (cf. Section~\ref{text:mpi}) of 62\%, which is calculated based on the average client model after round 1 with a MSE of 0.130.
Interestingly, the MSE of the global model is below the MSE of 0.064 achieved by the average client models.
This indicates that the individual training of the clients on their individual data sets leads to overfitting, while the global model balances this out and generalizes better.

Another interesting aspect is the comparison to the \emph{centralized} model, which represents the hypothetical, but practically infeasible case of training on the entire dataset.
Figure~\ref{fig:random_all} shows the performance for the centralized model (in black and dashed), which does not depend on the communication rounds as it is just trained once without a \FederatedLearning setup.
The centralized model exhibits a test MSE of 0.046 and a test MAE of 0.11, cf.~Figure~\ref{fig:random_all}, which is on par with the GNN developed by Sanchez Medina et al.~\cite{SanchezMedina.2023}.
Thus, the global model obtained through \FederatedLearning, exhibiting an MSE of 0.050, achieves a very similar prediction quality.
In the even-random scenario, \FederatedLearning therefore enables almost full utilization of the molecular property information from the individual clients without the need to share proprietary data, demonstrating great potential for collaborative model training.

Next, we consider the results for the uneven-scaffold scenario, where clients only share a part of the model.
The corresponding key question is whether the obtained benefits through \FederatedLearning can also be translated to more challenging and practical settings that come with data heterogeneity.

\begin{figure}[b!]
	\centering
	\begin{subfigure}[b]{0.5\textwidth}
		\centering
		\includegraphics[width=\linewidth]{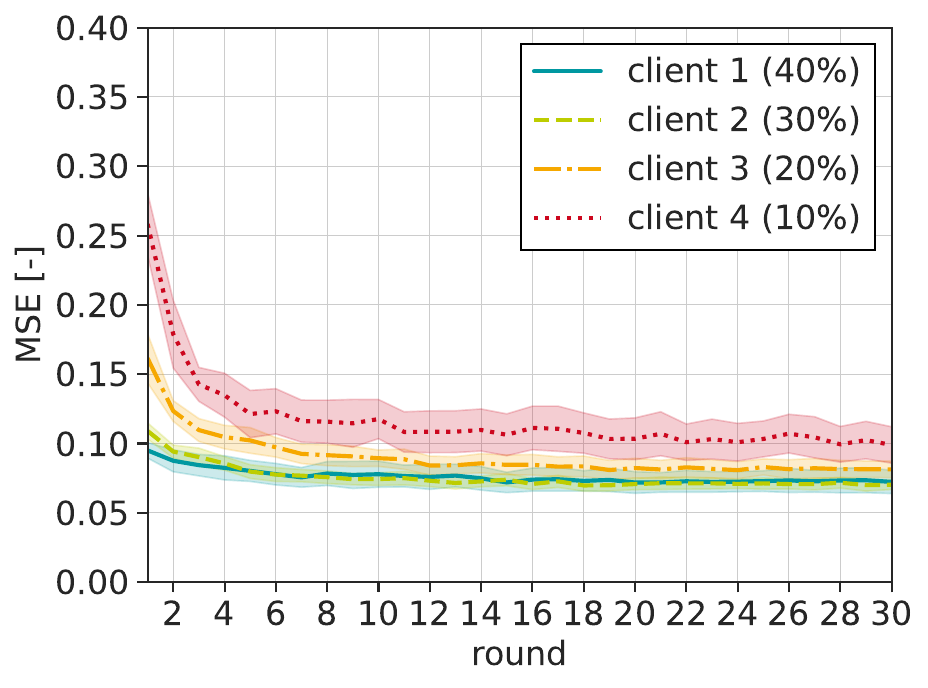}
		\caption{}
		\label{fig:scaffold_clients}
	\end{subfigure}\hfill
	\begin{subfigure}[b]{0.5\textwidth}
		\centering
		\includegraphics[width=\linewidth]{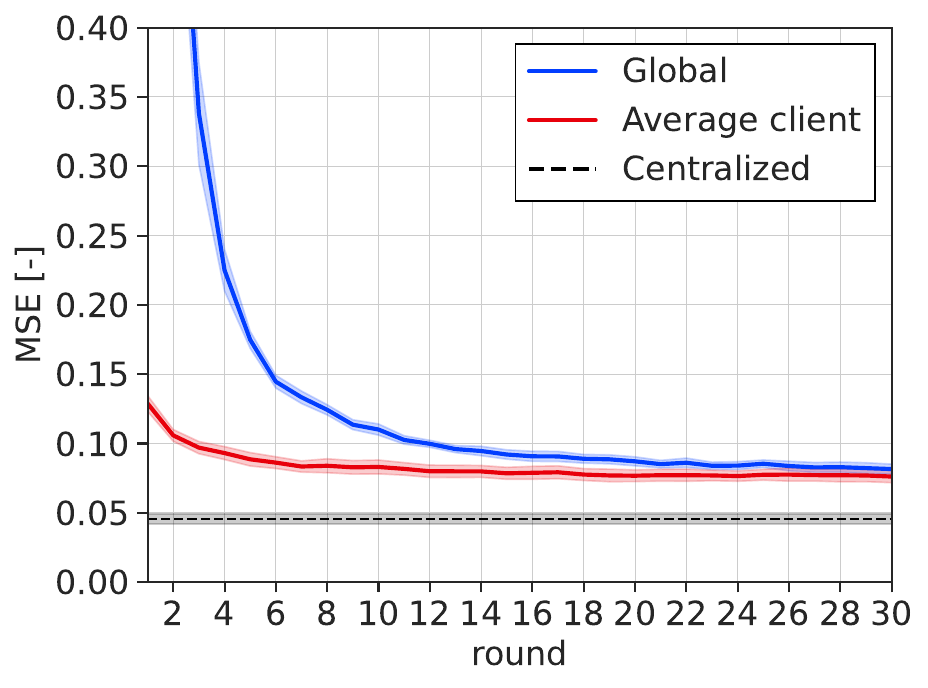}
		\caption{}
		\label{fig:scaffold_all}
	\end{subfigure}
	\caption{The test mean squared error (MSE) values for the \textbf{uneven-scaffold (partial model shared)} scenario: \textbf{(a)} results for the respective model of the \emph{individual clients} with different data set fractions; \textbf{(b)} aggregated results for the \emph{global} model that is obtained after each round through aggregation of the individual clients' models, the \emph{average client}, i.e., the averaged performances of the individual clients' models from (a), and \emph{centralized} model that is trained on the whole data set data and serves as a baseline.}
	\label{fig:scaffold}
\end{figure}

Considering the test MSE of the individual client models before aggregation, shown in Figure~\ref{fig:scaffold}, we find similar results to the even-random split: the test MSE values decrease over the communication rounds; all clients benefit from \FederatedLearning.
As expected, client 1 with the largest share of the data (40\%) shows the smallest decrease in MSE (0.095 to 0.072), whereas client 4 with the smallest share of data (10\%) obtains the largest improvement of the MSE (0.258 to 0.099).
The different performance increases highlight the challenge of generating incentives and valuating contributions to a \FederatedLearning collaboration, cf. Section~\ref{text:FLCE_challenges}.

We further investigate the global model performance in Figure~\ref{fig:scaffold}.
It is important to note that, in the uneven-scaffold scenario, the global model refers to a partly shared ML model between the clients, cf.~\ref{text:gnn:data}.
Hence for model evaluation, the globally shared part, i.e., the GNN layers, are combined with the private MLP heads of the clients, thereby representing the performance the clients obtain by using the received global model parameters; we refer to the Flower framework for implementation details~\cite{Beutel.2020}.
Again, we find a drastic decrease in the test MSE in the first communication rounds, followed by rather slow decreases in later rounds.
In contrast to the even-random case, we find that the global model and average client models are on a similar level after 30 rounds, with test MSE values of 0.082 and 0.076, respectively.
We suspect that both the increased data heterogeneity of the individual clients and the fact that only a part of the model is shared make it more difficult to balance the individual model contributions.  
This is also reflected in the comparison to the centralized model, which is the same model as in the even-random scenario, and thus exhibits a test MSE of 0.046.
Still, the final global model outperforms models that are trained by each client individually without \FederatedLearning, which have an average MSE of 0.156, resulting in a RIPtoP of 47\%.
In fact, this includes both the average client with a test MSE of 0.129 and the data-richest client (i.e., client 1) with a test MSE of 0.095.
Overall, our results show that \FederatedLearning provides all clients with an enhanced model performance, even in more challenging settings of data heterogeneity.

\subsection{Case Study II: Federated Learning for System Identification of Chemical Processes}\label{text:mpc}

\noindent As a second case study, we explore \FederatedLearning for system identification of chemical processes, which is an active field in \ChemicalEngineering and process control~\cite{Wu.2024b}.
While transfer learning has been proposed to reduce the amount of data needed to fit data-driven models for system identification and modeling~\cite{xiao2023modeling, Xiao.2024, niu2022deep}, we propose to utilize \FederatedLearning as a privacy-preserving alternative.
Specifically, we investigate the training of a data-driven model through \FederatedLearning to capture the dynamics of a high-purity distillation column.

\subsubsection{Model}\label{preliminaries_mpc:MIMO}

\noindent We use a multi-input multi-output~(MIMO) Wiener-type model that was developed for system identification and model reduction by Schulze \& Mitsos~\cite{Schulze.2022a}. 
Based on Koopman theory, the model architecture consists of a linear state-space model sandwiched by an autoencoder, as schematically illustrated in Figure~\ref{fig:case_studies_overview}.
That is, the physical states $\mathbf{x}(t)$ are first mapped to a lower-dimensional space, the latent space $\mathbf{z}(t)$, by a neural network, referred to as encoder.
Then, the linear state-space model is applied to capture the system dynamics.
Subsequently, the variables in the latent space are decoded back to the physical states by a second neural network, referred to as decoder.
The model is trained on full-state trajectories of length $p$ that are simulated responses to steps in the system inputs $\mathbf{u}(t)$. 
For more details on the model architecture, the training process, and the loss function, we refer to the original work~\cite{Schulze.2022a}.

As a chemical process, we consider a distillation column, as illustrated in Figure~\ref{fig:case_studies_overview}.
Specifically, we choose the methanol-propanol high-purity distillation column model studied in~\cite{Schulze.2022a, Jacobsen.1991, Pearson.2000} and reimplemented in~\cite{Schulze.2022a}. 
The column consists of eight trays, a reboiler, and a condenser. 
As in~\cite{Schulze.2022a}, the feed composition $x_F(t)$ and the liquid flow rate $L(t)$ are inputs of the system $\mathbf{u}(t)$ that vary over time.

\subsubsection{Data Set \& Federated Learning Scenario}

\begin{figure}[bt]
	\centering
	\includegraphics[width=0.5\linewidth]{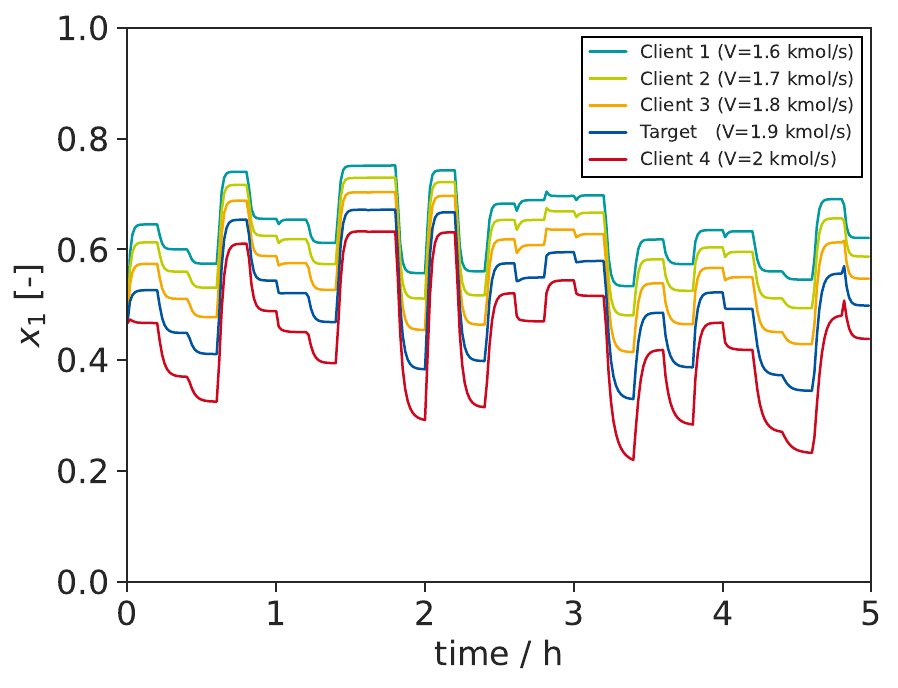}
	\caption{The evolution of the composition in the condenser, $x_1(t)$, in the test sets for different values of the vapor flow rate~$V$. All simulations start from the same initial state and are subject to the same input signals $\mathbf{u}(t)$.}
	\label{fig:trajectories}
\end{figure}

\noindent We simulate five different column models and create heterogeneity with different values of the vapor flow rate $V(t)$, cf. Figure~\ref{fig:case_studies_overview}. 
Specifically, we generate data sets of 192 trajectories assuming a constant $V$ $\in \{1.6, 1.7, 1.8, 1.9, 2.0\}$ kmol/s.
Exemplary trajectories for different vapor flow rates are shown in Figure~\ref{fig:trajectories}. 

For testing \FederatedLearning, we consider a scenario in which five different companies operate the same distillation column but each at different vapor flow rates $V \in$ \{1.6, 1.7, 1.8, 2.0\} kmol/s:
Four companies have extensively collected system response data, i.e., a \emph{rich} data set of 192 trajectories, which is the same amount of data as used by Schulze \& Mitsos~\cite{Schulze.2022a}.
A fifth company, the \emph{target} company, which operates the same distillation column at a different vapor flow rate ($V = 1.9$ kmol/s), has collected only a \emph{scarce} data set with 2 trajectories so far.
We are thus interested in whether \FederatedLearning provides a benefit in training a MIMO Wiener-type model for system identification for the company with fewer data.
For this, we assume that the companies apply \FederatedLearning with FedAvg as the aggregation algorithm and share the whole model.

We compare the federated learning scenario to the fifth company (i)~training a model only on their \textbf{private scarce data} set of 2 trajectories and (ii)~training a model on a \textbf{full data} set of 192 trajectories for $V = 1.9$ kmol/s, i.e., the hypothetical scenario of having the same amount of data available for training as the other four companies.

All models are evaluated in test sets of five hours in length, see Figure~\ref{fig:trajectories}.
For testing, we pass only the initial state $\mathbf{x}_0$ and the inputs $\mathbf{u}(t)$ to the model.
Then, we perform a multistep prediction for the entire five hours and compute the multi-step prediction MSE, cf.~\cite{Schulze.2022a}. 

\subsubsection{Implementation \& Hyperparameters}

\noindent For the MIMO Wiener-type model, we use the implementation by Schulze \& Mitsos~\cite{Schulze.2022a}, which is in Python and uses TensorFlow~\cite{TensorFlowDevelopers.2024}.
The mechanistic column models are implemented in Pyomo~\cite{Hart.2011, Bynum.2021} and simulated using CasADi~\cite{Andersson.2019}, analogously to~\cite{Schulze.2022a}.

We scale all data linearly to the range between zero and one based on the maximum and minimum of each state in the source data, such that we apply the same scaling to all column models. 
For the autoencoder, we use the same hyperparameters as in~\cite{Schulze.2022a} and map the ten physical states to a latent space of dimension two.
We repeat all computational runs with ten different seeds. 

\begin{figure}[b]
	\begin{center}
		\includegraphics[ width=0.65\textwidth]{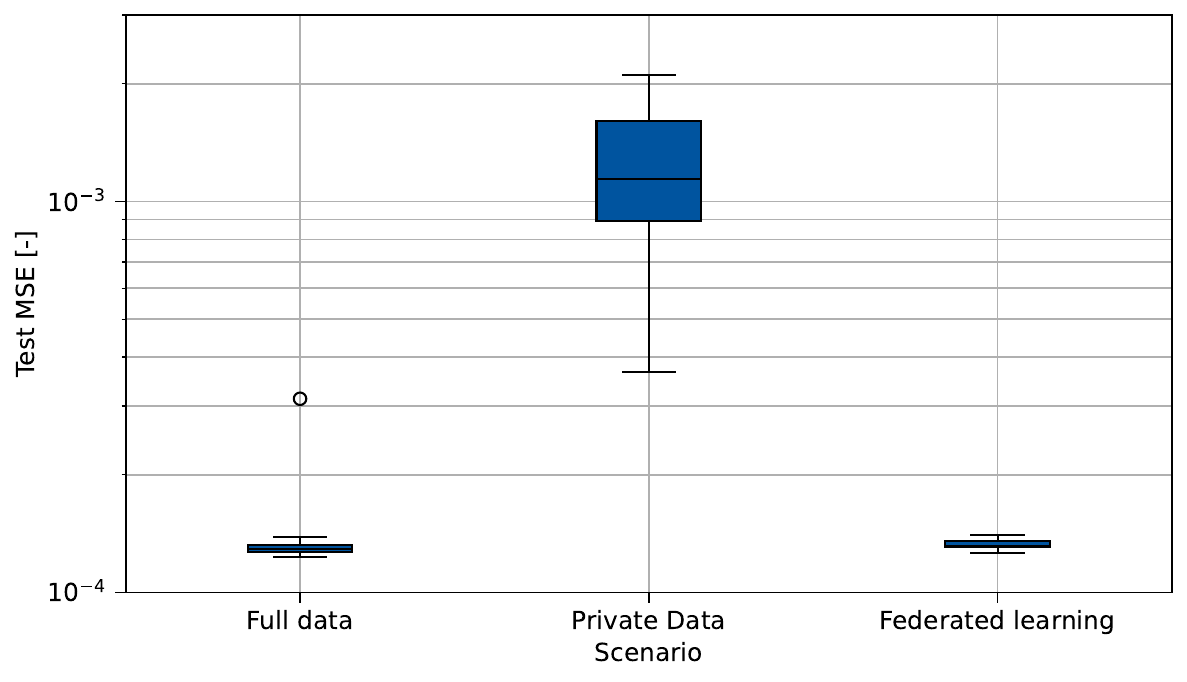}
		\caption[Test mean squarred error (MSE) for two different data availability scenarios and federated learning.]{Test mean squarred error (MSE) for two data availability scenarios, full data set with 192 trajectories and private data set with 2 trajectories, in comparison to federated learning in collaboration with multiple companies. Each boxplot is based on 10 runs with different seeds.}
		\label{fig:boxplot_SysID_paper}
	\end{center}
\end{figure}

\subsubsection{Results \& Discussion}\label{preliminaries_mpc:results}

\noindent We show the results for \FederatedLearning in comparison to the two baselines, i.e., training on the private scarce data set and on the full data set, in Figure~\ref{fig:boxplot_SysID_paper}.
The test MSE for the full data set baseline is $\num{1.47e-4}$ (and $\num{1.16e-4}$ excluding the outlier, see Figure~\ref{fig:boxplot_SysID_paper}), which is similar to the errors reported in~\cite{Schulze.2022a}.
The private data set baseline results in an MSE of $\num{1.44e-3}$, hence a significant increase of about one order of magnitude compared to the full data set case.
Thus, the small amount of data available for the target company does not suffice to train a model with similar accuracy as the other companies with more historical system response data.
We note that such an accuracy might still suffice for practical application in process control; yet, we want to investigate if this accuracy can be improved with \FederatedLearning.

Considering the \FederatedLearning scenario, we observe an MSE of $\num{1.33e-4}$, which corresponds to a RIPtoP of 91\%, cf. Figure~\ref{fig:boxplot_SysID_paper}.
\FederatedLearning thus enables the target company to decrease the error by an order of magnitude to a level similar to the other companies, without having to collect any additional process data.
Therefore, \FederatedLearning enables leveraging knowledge from similar source processes without exchange of underlying process data, as for example required in transfer learning~\cite{xiao2023modeling, niu2022deep}.

Notably, in our case study, the system responses for different vapor flow rates show similar dynamics, as can be seen in Figure~\ref{fig:trajectories}.
Future work could thus investigate whether \FederatedLearning can also provide accuracy improvements for more heterogeneous dynamics.
It would also be interesting to test \FederatedLearning for physics-informed neural networks (PINNs)~\cite{Velioglu.2025, Chen.2024} and transfer learning in process control applications~\cite{ArceMunoz.2024, Xiao.2024}.
In summary, \FederatedLearning shows promise for system identification, with many scenarios yet to be explored.

\section{Conclusion}\label{sec:Conclusion}

\noindent \FederatedLearning allows to jointly advance ML models without sharing data and is thus highly interesting for the chemical industry, where companies store large amounts of proprietary chemical and process data.
We identify numerous promising applications of \FederatedLearning in \ChemicalEngineering, including digitization, modeling, and design methods from the molecular up to the process scale.
We showcase in two such applications, namely mixture property prediction and system identification, that chemical companies can use \FederatedLearning in collaborations to greatly improve their ML models without the need to share proprietary data, making it highly promising for practical applications.
We anticipate and advocate for further \FederatedLearning applications in \ChemicalEngineering, e.g., in process flowsheet digitization and anomaly detection.

It will be of major importance to define benchmark data sets, raise incentives, elucidate data heterogeneity, and discuss potential risk through adversarial attacks in strong collaboration of academia, database organization, and chemical companies. 
Pilot projects with practical applications of \FederatedLearning will further elucidate challenges of data and collaboration within the chemical industry -- with the goal of addressing data scarcity and advancing ML in \ChemicalEngineering.

\section*{Data availability}
\noindent We provide the data and source code for the two presented case studies as open source in the following GitLab repositories:
\begin{itemize}
	\item Activity coefficient prediction: \url{https://git.rwth-aachen.de/avt-svt/public/gmolprop}
	\item System identification: \url{https://git.rwth-aachen.de/avt-svt/public/koopman-wiener-fl}
\end{itemize}

\section*{Acknowledgments}

\noindent This project was partly funded by the Werner Siemens Foundation within the WSS project of the century ``catalaix''. 
This work was also performed as part of the Helmholtz School for Data Science in Life, Earth and Energy (HDS-LEE). 
Simulations were performed with computing resources granted by RWTH Aachen University under project ``thes1813''. 

We sincerely thank Alexander Mitsos for his valuable support and feedback within this project.

\section*{Authors contributions}
\noindent \textbf{Jan G. Rittig}: Conceptualization, Formal analysis, Funding acquisition, Methodology, Supervision, Visualization, Writing - Original draft.\\
\textbf{Clemens Kortmann}: Conceptualization, Data curation, Formal analysis, Methodology, Software, Visualization, Writing - Original draft.

\section*{Conflict of interest}
There are no conflicts to declare.
\label{sec:others}

\bibliographystyle{unsrt}  
\bibliography{references}  %%% Remove comment to use the external .bib file (using bibtex).

\begin{thebibliography}{100}

\bibitem{Schweidtmann.2021}
Artur~M. Schweidtmann, Erik Esche, Asja Fischer, Marius Kloft, Jens-Uwe Repke,
  Sebastian Sager, and Alexander Mitsos.
\newblock Machine learning in chemical engineering: A perspective.
\newblock {\em Chemie Ingenieur Technik}, 93(12):2029--2039, 2021.

\bibitem{StriethKalthoff.2020}
Felix Strieth-Kalthoff, Frederik Sandfort, Marwin H.~S. Segler, and Frank
  Glorius.
\newblock Machine learning the ropes: principles, applications and directions
  in synthetic chemistry.
\newblock {\em Chemical Society reviews}, 49(17):6154--6168, 2020.

\bibitem{Daoutidis.2024}
Prodromos Daoutidis, Jay~H. Lee, Srinivas Rangarajan, Leo Chiang, Bhushan
  Gopaluni, Artur~M. Schweidtmann, Iiro Harjunkoski, Mehmet Mercang{\"o}z, Ali
  Mesbah, Fani Boukouvala, Fernando~V. Lima, Antonio {Del Rio Chanona}, and
  Christos Georgakis.
\newblock Machine learning in process systems engineering: Challenges and
  opportunities.
\newblock {\em Computers {\&} Chemical Engineering}, 181:108523, 2024.

\bibitem{Cheng.2024}
Austin Cheng, Marta Skreta, Cher~Tian Ser, Andres Guzman-Cordero, Luca Thiede,
  Andreas Burger, Sergio Pablo-Garc{\'i}a, Abdulrahman Aldossary, Shi~Xuan
  Leong, Felix Strieth-Kalthoff, and Alan Aspuru-Guzik.
\newblock How to do impactful research in artificial intelligence for chemistry
  and materials science.
\newblock {\em Faraday Discussions}, 2024.

\bibitem{hestness2017deep}
Joel Hestness, Sharan Narang, Newsha Ardalani, Gregory Diamos, Heewoo Jun,
  Hassan Kianinejad, Md~Mostofa~Ali Patwary, Yang Yang, and Yanqi Zhou.
\newblock Deep learning scaling is predictable, empirically.
\newblock {\em arXiv preprint arXiv:1712.00409}, 2017.

\bibitem{kaplan2020scaling}
Jared Kaplan, Sam McCandlish, Tom Henighan, Tom~B Brown, Benjamin Chess, Rewon
  Child, Scott Gray, Alec Radford, Jeffrey Wu, and Dario Amodei.
\newblock Scaling laws for neural language models.
\newblock {\em arXiv preprint arXiv:2001.08361}, 2020.

\bibitem{hoffmann2022training}
Jordan Hoffmann, Sebastian Borgeaud, Arthur Mensch, Elena Buchatskaya, Trevor
  Cai, Eliza Rutherford, Diego de~Las Casas, Lisa~Anne Hendricks, Johannes
  Welbl, Aidan Clark, et~al.
\newblock Training compute-optimal large language models.
\newblock {\em arXiv preprint arXiv:2203.15556}, 2022.

\bibitem{zhai2022scaling}
Xiaohua Zhai, Alexander Kolesnikov, Neil Houlsby, and Lucas Beyer.
\newblock Scaling vision transformers.
\newblock In {\em Proceedings of the IEEE/CVF conference on computer vision and
  pattern recognition}, pages 12104--12113, 2022.

\bibitem{qin2019advances}
S~Joe Qin and Leo~H Chiang.
\newblock Advances and opportunities in machine learning for process data
  analytics.
\newblock {\em Computers \& Chemical Engineering}, 126:465--473, 2019.

\bibitem{thebelt2022maximizing}
Alexander Thebelt, Johannes Wiebe, Jan Kronqvist, Calvin Tsay, and Ruth
  Misener.
\newblock Maximizing information from chemical engineering data sets:
  Applications to machine learning.
\newblock {\em Chemical Engineering Science}, 252:117469, 2022.

\bibitem{Zhu.2022}
Wei Zhu, Jiebo Luo, and Andrew~D. White.
\newblock Federated learning of molecular properties with graph neural networks
  in a heterogeneous setting.
\newblock {\em Patterns}, 3(6):100521, 2022.

\bibitem{Dutta.2024}
Siddhant Dutta, Iago Leal~de Freitas, Pedro Maciel~Xavier, Claudio Miceli~de
  Farias, and David~E Bernal~Neira.
\newblock Federated learning in chemical engineering: A tutorial on a framework
  for privacy-preserving collaboration across distributed data sources.
\newblock {\em Industrial \& Engineering Chemistry Research}, 2024.

\bibitem{geballe2010sampl2}
Matthew~T Geballe, A~Geoffrey Skillman, Anthony Nicholls, J~Peter Guthrie, and
  Peter~J Taylor.
\newblock The sampl2 blind prediction challenge: introduction and overview.
\newblock {\em Journal of computer-aided molecular design}, 24:259--279, 2010.

\bibitem{mobley2014freesolv}
David~L Mobley and J~Peter Guthrie.
\newblock Freesolv: a database of experimental and calculated hydration free
  energies, with input files.
\newblock {\em Journal of computer-aided molecular design}, 28:711--720, 2014.

\bibitem{qm9}
Raghunathan Ramakrishnan, Pavlo~O. Dral, Matthias Rupp, and O.~Anatole von
  Lilienfeld.
\newblock Quantum chemistry structures and properties of 134 kilo molecules.
\newblock {\em Scientific Data}, 1(1):140022, Aug 2014.

\bibitem{felton2021summit}
Kobi~C Felton, Jan~G Rittig, and Alexei~A Lapkin.
\newblock Summit: benchmarking machine learning methods for reaction
  optimisation.
\newblock {\em Chemistry-Methods}, 1(2):116--122, 2021.

\bibitem{wigh2024orderly}
Daniel~S Wigh, Joe Arrowsmith, Alexander Pomberger, Kobi~C Felton, and Alexei~A
  Lapkin.
\newblock Orderly: data sets and benchmarks for chemical reaction data.
\newblock {\em Journal of Chemical Information and Modeling}, 64(9):3790--3798,
  2024.

\bibitem{chiang2012fault}
Leo~H Chiang, Evan~L Russell, and Richard~D Braatz.
\newblock {\em Fault detection and diagnosis in industrial systems}.
\newblock Springer Science \& Business Media, 2012.

\bibitem{bradley2022perspectives}
William Bradley, Jinhyeun Kim, Zachary Kilwein, Logan Blakely, Michael
  Eydenberg, Jordan Jalvin, Carl Laird, and Fani Boukouvala.
\newblock Perspectives on the integration between first-principles and
  data-driven modeling.
\newblock {\em Computers \& Chemical Engineering}, 166:107898, 2022.

\bibitem{schweidtmann2024review}
Artur~M Schweidtmann, Dongda Zhang, and Moritz von Stosch.
\newblock A review and perspective on hybrid modeling methodologies.
\newblock {\em Digital Chemical Engineering}, 10:100136, 2024.

\bibitem{karniadakis2021physics}
George~Em Karniadakis, Ioannis~G Kevrekidis, Lu~Lu, Paris Perdikaris, Sifan
  Wang, and Liu Yang.
\newblock Physics-informed machine learning.
\newblock {\em Nature Reviews Physics}, 3(6):422--440, 2021.

\bibitem{boiko2023autonomous}
Daniil~A Boiko, Robert MacKnight, Ben Kline, and Gabe Gomes.
\newblock Autonomous chemical research with large language models.
\newblock {\em Nature}, 624(7992):570--578, 2023.

\bibitem{m2024augmenting}
Andres M.~Bran, Sam Cox, Oliver Schilter, Carlo Baldassari, Andrew~D White, and
  Philippe Schwaller.
\newblock Augmenting large language models with chemistry tools.
\newblock {\em Nature Machine Intelligence}, 6(5):525--535, 2024.

\bibitem{vermeire2021transfer}
Florence~H Vermeire and William~H Green.
\newblock Transfer learning for solvation free energies: From quantum chemistry
  to experiments.
\newblock {\em Chemical Engineering Journal}, 418:129307, 2021.

\bibitem{wu2020fault}
Hao Wu and Jinsong Zhao.
\newblock Fault detection and diagnosis based on transfer learning for
  multimode chemical processes.
\newblock {\em Computers \& Chemical Engineering}, 135:106731, 2020.

\bibitem{li2020transfer}
Weijun Li, Sai Gu, Xiangping Zhang, and Tao Chen.
\newblock Transfer learning for process fault diagnosis: Knowledge transfer
  from simulation to physical processes.
\newblock {\em Computers \& Chemical Engineering}, 139:106904, 2020.

\bibitem{savage2023multi}
Tom Savage, Nausheen Basha, Jonathan McDonough, Omar~K Matar, and
  Ehecatl~Antonio del Rio~Chanona.
\newblock Multi-fidelity data-driven design and analysis of reactor and tube
  simulations.
\newblock {\em Computers \& Chemical Engineering}, 179:108410, 2023.

\bibitem{nevolianis2024multi}
Thomas Nevolianis, Jan~Gerald Rittig, Alexander Mitsos, and Kai Leonhard.
\newblock Multi-fidelity graph neural networks for predicting toluene/water
  partition coefficients.
\newblock {\em ChemRxiv preprint 10.26434/chemrxiv-2024-3t818}, 2024.

\bibitem{McMahan.2017}
H.~Brendan McMahan, Eider Moore, Daniel Ramage, Seth Hampson, and Blaise
  {Ag{\"u}era y Arcas}.
\newblock Communication-efficient learning of deep networks from decentralized
  data.
\newblock {\em Artificial Intelligence and Statistics}, pages 1273--1282, 2017.

\bibitem{Kairouz.2019}
Peter Kairouz, H.~Brendan McMahan, Brendan Avent, Aurelien Bellet, and Sen
  Zhao.
\newblock Advances and open problems in federated learning.
\newblock Foundations and trends{\circledR} in machine learning 14.1--2 (2021):
  1-210., 2021.

\bibitem{Daly.2024}
Katharine Daly, Hubert Eichner, Peter Kairouz, H.~Brendan McMahan, Daniel
  Ramage, and Zheng Xu.
\newblock Federated learning in practice: Reflections and projections.
\newblock {\em arXiv Preprint arXiv:2410.08892v1}, 2024.

\bibitem{Liu.2024a}
Bingyan Liu, Nuoyan Lv, Yuanchun Guo, and Yawen Li.
\newblock Recent advances on federated learning: A systematic survey.
\newblock {\em Neurocomputing}, 597:128019, 2024.

\bibitem{Wen.2023}
Jie Wen, Zhixia Zhang, Yang Lan, Zhihua Cui, Jianghui Cai, and Wensheng Zhang.
\newblock A survey on federated learning: challenges and applications.
\newblock {\em International Journal of Machine Learning and Cybernetics},
  14(2):513--535, 2023.

\bibitem{Li.2024}
Shenghui Li, Fanghua Ye, Meng Fang, Jiaxu Zhao, Yun-Hin Chan, Edith C.~H Ngai,
  and Thiemo Voigt.
\newblock Synergizing foundation models and federated learning: A survey.
\newblock {\em arXiv Preprint arXiv:2406.12844v1}, 2024.

\bibitem{rieke2020future}
Nicola Rieke, Jonny Hancox, Wenqi Li, Fausto Milletari, Holger~R Roth, Shadi
  Albarqouni, Spyridon Bakas, Mathieu~N Galtier, Bennett~A Landman, Klaus
  Maier-Hein, et~al.
\newblock The future of digital health with federated learning.
\newblock {\em NPJ digital medicine}, 3(1):119, 2020.

\bibitem{Nguyen.2023b}
Dinh~C. Nguyen, Quoc-Viet Pham, Pubudu~N. Pathirana, Ming Ding, Aruna
  Seneviratne, Zihuai Lin, Octavia Dobre, and Won-Joo Hwang.
\newblock Federated learning for smart healthcare: A survey.
\newblock {\em ACM Computing Surveys}, 55(3):1--37, 2023.

\bibitem{Rauniyar.2024}
Ashish Rauniyar, Desta~Haileselassie Hagos, Debesh Jha, Jan~Erik
  H{\aa}keg{\aa}rd, Ulas Bagci, Danda~B. Rawat, and Vladimir Vlassov.
\newblock Federated learning for medical applications: A taxonomy, current
  trends, challenges, and future research directions.
\newblock {\em IEEE Internet of Things Journal}, 11(5):7374--7398, 2024.

\bibitem{Heyndrickx.2024}
Wouter Heyndrickx, Lewis Mervin, Tobias Morawietz, No{\'e} Sturm, Lukas
  Friedrich, Adam Zalewski, Anastasia Pentina, Lina Humbeck, Martijn Oldenhof,
  Ritsuya Niwayama, Peter Schmidtke, Nikolas Fechner, Jaak Simm, Adam Arany,
  Nicolas Drizard, Rama Jabal, Arina Afanasyeva, Regis Loeb, Shlok Verma, Simon
  Harnqvist, Matthew Holmes, Balazs Pejo, Maria Telenczuk, Nicholas Holway,
  Arne Dieckmann, Nicola Rieke, Friederike Zumsande, Djork-Arn{\'e} Clevert,
  Michael Krug, Christopher Luscombe, Darren Green, Peter Ertl, Peter Antal,
  David Marcus, Nicolas {Do Huu}, Hideyoshi Fuji, Stephen Pickett, Gergely Acs,
  Eric Boniface, Bernd Beck, Yax Sun, Arnaud Gohier, Friedrich Rippmann, Ola
  Engkvist, Andreas~H. G{\"o}ller, Yves Moreau, Mathieu~N. Galtier, Ansgar
  Schuffenhauer, and Hugo Ceulemans.
\newblock Melloddy: Cross-pharma federated learning at unprecedented scale
  unlocks benefits in {QSAR} without compromising proprietary information.
\newblock {\em Journal of chemical information and modeling}, 64(7):2331--2344,
  2024.

\bibitem{yang2018applied}
Timothy Yang, Galen Andrew, Hubert Eichner, Haicheng Sun, Wei Li, Nicholas
  Kong, Daniel Ramage, and Fran{\c{c}}oise Beaufays.
\newblock Applied federated learning: Improving google keyboard query
  suggestions.
\newblock {\em arXiv preprint arXiv:1812.02903}, 2018.

\bibitem{Sani.2024a}
Lorenzo Sani, Alex Iacob, Zeyu Cao, Bill Marino, Yan Gao, Tomas Paulik, Wanru
  Zhao, William~F. Shen, Preslav Aleksandrov, Xinchi Qiu, and Nicholas~D. Lane.
\newblock The future of large language model pre-training is federated.
\newblock {\em arXiv Preprint arXiv:2405.10853v3}, 2024.

\bibitem{lu2022federated}
Ming~Y Lu, Richard~J Chen, Dehan Kong, Jana Lipkova, Rajendra Singh, Drew~FK
  Williamson, Tiffany~Y Chen, and Faisal Mahmood.
\newblock Federated learning for computational pathology on gigapixel whole
  slide images.
\newblock {\em Medical image analysis}, 76:102298, 2022.

\bibitem{guan2024federated}
Hao Guan, Pew-Thian Yap, Andrea Bozoki, and Mingxia Liu.
\newblock Federated learning for medical image analysis: A survey.
\newblock {\em Pattern Recognition}, page 110424, 2024.

\bibitem{Heyndrickx.2023}
Wouter Heyndrickx, Adam Arany, Jaak Simm, Anastasia Pentina, No{\'e} Sturm,
  Lina Humbeck, Lewis Mervin, Adam Zalewski, Martijn Oldenhof, Peter Schmidtke,
  Lukas Friedrich, Regis Loeb, Arina Afanasyeva, Ansgar Schuffenhauer, Yves
  Moreau, and Hugo Ceulemans.
\newblock Conformal efficiency as a metric for comparative model assessment
  befitting federated learning.
\newblock {\em Artificial Intelligence in the Life Sciences}, 3:100070, 2023.

\bibitem{Oldenhof.2023}
Martijn Oldenhof, Gergely {\'A}cs, Bal{\'a}zs Pej{\'o}, Ansgar Schuffenhauer,
  Nicholas Holway, No{\'e} Sturm, Arne Dieckmann, Oliver Fortmeier, Eric
  Boniface, Cl{\'e}ment Mayer, Arnaud Gohier, Peter Schmidtke, Ritsuya
  Niwayama, Dieter Kopecky, Lewis Mervin, Prakash~Chandra Rathi, Lukas
  Friedrich, Andr{\'a}s Formanek, Peter Antal, Jordon Rahaman, Adam Zalewski,
  Wouter Heyndrickx, Ezron Oluoch, Manuel St{\"o}{\ss}el, Michal Van{\v{c}}o,
  David Endico, Fabien Gelus, Tha{\"i}s de~Boisfoss{\'e}, Adrien Darbier,
  Ashley Nicollet, Matthieu Blotti{\`e}re, Maria Telenczuk, Tien {van Nguyen},
  Thibaud Martinez, Camille Boillet, Kelvin Moutet, Alexandre Picosson,
  Aur{\'e}lien Gasser, Inal Djafar, Antoine Simon, {\'A}d{\'a}m Arany, Jaak
  Simm, Yves Moreau, Ola Engkvist, Hugo Ceulemans, Camille Marini, and Mathieu
  Galtier.
\newblock Industry-scale orchestrated federated learning for drug discovery.
\newblock {\em Proceedings of the AAAI Conference on Artificial Intelligence},
  37(13):15576--15584, 2023.

\bibitem{wu2022communication}
Chuhan Wu, Fangzhao Wu, Lingjuan Lyu, Yongfeng Huang, and Xing Xie.
\newblock Communication-efficient federated learning via knowledge
  distillation.
\newblock {\em Nature communications}, 13(1):2032, 2022.

\bibitem{Hanser.2025}
Thierry Hanser, Ernst Ahlberg, Alexander Amberg, Lennart~T. Anger, Chris
  Barber, Richard~J. Brennan, Alessandro Brigo, Annie Delaunois, Susanne
  Glowienke, Nigel Greene, Laura Johnston, Daniel Kuhn, Lara Kuhnke,
  Jean-Fran{\c{c}}ois Marchaland, Wolfgang Muster, Jeffrey Plante, Friedrich
  Rippmann, Yogesh Sabnis, Friedemann Schmidt, Ruud {van Deursen}, St{\'e}phane
  Werner, Angela White, Joerg Wichard, and Tomoya Yukawa.
\newblock Data-driven federated learning in drug discovery with knowledge
  distillation.
\newblock {\em Nature Machine Intelligence}, 7(3):423--436, 2025.

\bibitem{Hayer.2021}
Mathis Hayer.
\newblock Decentralized and incentivized federated learning for the chemical
  engineering domain.
\newblock Master's thesis, Tsinghua University, 2021.
\newblock This thesis is not available online.

\bibitem{Xu.2024}
Zeyuan Xu and Zhe Wu.
\newblock Privacy-preserving federated machine learning modeling and predictive
  control of heterogeneous nonlinear systems.
\newblock {\em Computers {\&} Chemical Engineering}, 187:108749, 2024.

\bibitem{Cheng.2022}
Xu~Cheng, Chendan Li, and Xiufeng Liu.
\newblock A review of federated learning in energy systems.
\newblock In {\em 2022 IEEE/IAS Industrial and Commercial Power System Asia
  (I{\&}CPS Asia)}. IEEE, 2022.

\bibitem{grataloup2024review}
Albin Grataloup, Stefan Jonas, and Angela Meyer.
\newblock A review of federated learning in renewable energy applications:
  Potential, challenges, and future directions.
\newblock {\em Energy and AI}, page 100375, 2024.

\bibitem{huang2022cross}
Chao Huang, Jianwei Huang, and Xin Liu.
\newblock Cross-silo federated learning: Challenges and opportunities.
\newblock {\em arXiv preprint arXiv:2206.12949}, 2022.

\bibitem{Zhou.2024}
Tailin Zhou, Zehong Lin, Jun Zhang, and Danny~H.K. Tsang.
\newblock Understanding and improving model averaging in federated learning on
  heterogeneous data.
\newblock {\em IEEE Transactions on Mobile Computing}, pages 1--16, 2024.

\bibitem{Yang.2024}
Shusen Yang, Fangyuan Zhao, Zihao Zhou, Liang Shi, Xuebin Ren, and Zongben Xu.
\newblock Review of mathematical optimization in federated learning.
\newblock {\em arXiv Preprint arXiv:2412.01630v1}, 2024.

\bibitem{Smith.2017}
Virginia Smith, Chao-Kai Chiang, Maziar Sanjabi, and Ameet~S. Talwalkar.
\newblock Federated multi-task learning.
\newblock In I.~Guyon, U.~Von Luxburg, S.~Bengio, H.~Wallach, R.~Fergus,
  S.~Vishwanathan, and R.~Garnett, editors, {\em Advances in Neural Information
  Processing Systems}, volume~30. {Curran Associates, Inc}, 2017.

\bibitem{Qi.2024}
Pian Qi, Diletta Chiaro, Antonella Guzzo, Michele Ianni, Giancarlo Fortino, and
  Francesco Piccialli.
\newblock Model aggregation techniques in federated learning: A comprehensive
  survey.
\newblock {\em Future Generation Computer Systems}, 150:272--293, 2024.

\bibitem{Dinh.2024}
Canh~T. Dinh, Tung~T. Vu, and Nguyen~H. Tran.
\newblock Chapter 7 - personalized federated learning: theory and open
  problems.
\newblock In Lam~M. Nguyen, editor, {\em Federated Learning}, pages 125--141.
  {Elsevier Science {\&} Technology}, San Diego, 2024.

\bibitem{shao2023survey}
Jiawei Shao, Zijian Li, Wenqiang Sun, Tailin Zhou, Yuchang Sun, Lumin Liu,
  Zehong Lin, Yuyi Mao, and Jun Zhang.
\newblock A survey of what to share in federated learning: Perspectives on
  model utility, privacy leakage, and communication efficiency.
\newblock {\em arXiv preprint arXiv:2307.10655}, 2023.

\bibitem{wang2022unreasonable}
Jianyu Wang, Rudrajit Das, Gauri Joshi, Satyen Kale, Zheng Xu, and Tong Zhang.
\newblock On the unreasonable effectiveness of federated averaging with
  heterogeneous data.
\newblock {\em arXiv preprint arXiv:2206.04723}, 2022.

\bibitem{Rittig.2024}
Jan~G. Rittig and Alexander Mitsos.
\newblock Thermodynamics-consistent graph neural networks.
\newblock {\em Arxiv Preprint arXiv:2407.18372}, 2024.

\bibitem{theisen2023digitization}
Maximilian~F Theisen, Kenji~Nishizaki Flores, Lukas~Schulze Balhorn, and
  Artur~M Schweidtmann.
\newblock Digitization of chemical process flow diagrams using deep
  convolutional neural networks.
\newblock {\em Digital chemical engineering}, 6:100072, 2023.

\bibitem{stops2023flowsheet}
Laura Stops, Roel Leenhouts, Qinghe Gao, and Artur~M Schweidtmann.
\newblock Flowsheet generation through hierarchical reinforcement learning and
  graph neural networks.
\newblock {\em AIChE Journal}, 69(1):e17938, 2023.

\bibitem{Ekramipooya.2024}
Ali Ekramipooya, Mehrdad Boroushaki, and Davood Rashtchian.
\newblock Predicting possible recommendations related to causes and
  consequences in the hazop study worksheet using natural language processing
  and machine learning: Bert, clustering, and classification.
\newblock {\em Journal of Loss Prevention in the Process Industries},
  89:105310, 2024.

\bibitem{kang2019digitization}
Sung-O Kang, Eul-Bum Lee, and Hum-Kyung Baek.
\newblock A digitization and conversion tool for imaged drawings to intelligent
  piping and instrumentation diagrams {(P\&ID)}.
\newblock {\em Energies}, 12(13):2593, 2019.

\bibitem{Sturmer.2024}
Jan~Marius St{\"u}rmer, Marius Graumann, and Tobias Koch.
\newblock Transforming engineering diagrams: A novel approach for {P}{\&}{ID}
  digitization using transformers.
\newblock {\em arXiv Preprint arXiv:2411.13929v1}, 2024.

\bibitem{Balhorn.2022}
Lukas~Schulze Balhorn, Qinghe Gao, Dominik Goldstein, and Artur~M.
  Schweidtmann.
\newblock Flowsheet recognition using deep convolutional neural networks.
\newblock In Yoshiyuki Yamashita and Manabu Kano, editors, {\em 14th
  International Symposium on Process Systems Engineering}, volume~49 of {\em
  Computer-aided chemical engineering}, pages 1567--1572. Elsevier, Amsterdam
  and Boston and Heidelberg and London, 2022.

\bibitem{bhosekar2018advances}
Atharv Bhosekar and Marianthi Ierapetritou.
\newblock Advances in surrogate based modeling, feasibility analysis, and
  optimization: A review.
\newblock {\em Computers \& Chemical Engineering}, 108:250--267, 2018.

\bibitem{schweidtmann2019deterministic}
Artur~M Schweidtmann and Alexander Mitsos.
\newblock Deterministic global optimization with artificial neural networks
  embedded.
\newblock {\em Journal of Optimization Theory and Applications},
  180(3):925--948, 2019.

\bibitem{mcbride2019overview}
Kevin McBride and Kai Sundmacher.
\newblock Overview of surrogate modeling in chemical process engineering.
\newblock {\em Chemie Ingenieur Technik}, 91(3):228--239, 2019.

\bibitem{esche2022architectures}
Erik Esche, Joris Weigert, Gerardo~Brand Rihm, Jan G{\"o}bel, and Jens-Uwe
  Repke.
\newblock Architectures for neural networks as surrogates for dynamic systems
  in chemical engineering.
\newblock {\em Chemical Engineering Research and Design}, 177:184--199, 2022.

\bibitem{misener2023formulating}
Ruth Misener and Lorenz Biegler.
\newblock Formulating data-driven surrogate models for process optimization.
\newblock {\em Computers \& Chemical Engineering}, 179:108411, 2023.

\bibitem{sansana2021recent}
Joel Sansana, Mark~N Joswiak, Ivan Castillo, Zhenyu Wang, Ricardo Rendall,
  Leo~H Chiang, and Marco~S Reis.
\newblock Recent trends on hybrid modeling for industry 4.0.
\newblock {\em Computers \& Chemical Engineering}, 151:107365, 2021.

\bibitem{meuwly2021machine}
Markus Meuwly.
\newblock Machine learning for chemical reactions.
\newblock {\em Chemical Reviews}, 121(16):10218--10239, 2021.

\bibitem{margraf2023exploring}
Johannes~T Margraf, Hyunwook Jung, Christoph Scheurer, and Karsten Reuter.
\newblock Exploring catalytic reaction networks with machine learning.
\newblock {\em Nature Catalysis}, 6(2):112--121, 2023.

\bibitem{weber2021chemical}
Jana~M Weber, Zhen Guo, Chonghuan Zhang, Artur~M Schweidtmann, and Alexei~A
  Lapkin.
\newblock Chemical data intelligence for sustainable chemistry.
\newblock {\em Chemical Society Reviews}, 50(21):12013--12036, 2021.

\bibitem{jorner2021machine}
Kjell Jorner, Tore Brinck, Per-Ola Norrby, and David Buttar.
\newblock Machine learning meets mechanistic modelling for accurate prediction
  of experimental activation energies.
\newblock {\em Chemical Science}, 12(3):1163--1175, 2021.

\bibitem{chung2024machine}
Yunsie Chung and William~H Green.
\newblock Machine learning from quantum chemistry to predict experimental
  solvent effects on reaction rates.
\newblock {\em Chemical Science}, 15(7):2410--2424, 2024.

\bibitem{quaglio2020artificial}
Marco Quaglio, Louise Roberts, Mohd Safarizal~Bin Jaapar, Eric~S Fraga, Vivek
  Dua, and Federico Galvanin.
\newblock An artificial neural network approach to recognise kinetic models
  from experimental data.
\newblock {\em Computers \& Chemical Engineering}, 135:106759, 2020.

\bibitem{schweidtmann2024generative}
Artur~M Schweidtmann.
\newblock Generative artificial intelligence in chemical engineering.
\newblock {\em Nature Chemical Engineering}, 1(3):193--193, 2024.

\bibitem{gottl2022automated}
Quirin G{\"o}ttl, Dominik~G Grimm, and Jakob Burger.
\newblock Automated synthesis of steady-state continuous processes using
  reinforcement learning.
\newblock {\em Frontiers of Chemical Science and Engineering}, pages 1--15,
  2022.

\bibitem{Vogel.2023}
Gabriel Vogel, Lukas {Schulze Balhorn}, and Artur~M. Schweidtmann.
\newblock Learning from flowsheets: A generative transformer model for
  autocompletion of flowsheets.
\newblock {\em Computers {\&} Chemical Engineering}, 171:108162, 2023.

\bibitem{mann2024esfiles}
Vipul Mann, Mauricio Sales-Cruz, Rafiqul Gani, and Venkat Venkatasubramanian.
\newblock esfiles: Intelligent process flowsheet synthesis using process
  knowledge, symbolic ai, and machine learning.
\newblock {\em Computers \& Chemical Engineering}, 181:108505, 2024.

\bibitem{Schulze-Balhorn.2024}
Lukas {Schulze Balhorn}, Marc Caballero, and Artur~M. Schweidtmann.
\newblock Toward autocorrection of chemical process flowsheets using large
  language models.
\newblock In Flavio Manenti and Gintaras~V. Reklaitis, editors, {\em Computer
  Aided Chemical Engineering : 34 European Symposium on Computer Aided Process
  Engineering / 15 International Symposium on Process Systems Engineering},
  volume~53, pages 3109--3114. Elsevier, 2024.

\bibitem{balhorn2025graph}
Lukas~Schulze Balhorn, Kevin Degens, and Artur~M Schweidtmann.
\newblock Graph-to-sfiles: Control structure prediction from process topologies
  using generative artificial intelligence.
\newblock {\em Computers \& Chemical Engineering}, page 109121, 2025.

\bibitem{gao2024deep}
Qinghe Gao and Artur~M Schweidtmann.
\newblock Deep reinforcement learning for process design: Review and
  perspective.
\newblock {\em Current Opinion in Chemical Engineering}, 44:101012, 2024.

\bibitem{gottl2025deep}
Quirin G{\"o}ttl, Jonathan Pirnay, Jakob Burger, and Dominik~G Grimm.
\newblock Deep reinforcement learning enables conceptual design of processes
  for separating azeotropic mixtures without prior knowledge.
\newblock {\em Computers \& Chemical Engineering}, 194:108975, 2025.

\bibitem{tang2022data}
Wentao Tang and Prodromos Daoutidis.
\newblock Data-driven control: Overview and perspectives.
\newblock In {\em 2022 American Control Conference (ACC)}, pages 1048--1064.
  IEEE, 2022.

\bibitem{chiuso2019system}
Alessandro Chiuso and Gianluigi Pillonetto.
\newblock System identification: A machine learning perspective.
\newblock {\em Annual Review of Control, Robotics, and Autonomous Systems},
  2(1):281--304, 2019.

\bibitem{ahmed2025comparative}
Akhil Ahmed, Ehecatl~Antonio del Rio-Chanona, and Mehmet Mercangöz.
\newblock Comparative study of machine learning and system identification for
  process systems engineering dynamics.
\newblock {\em Industrial \& Engineering Chemistry Research}, 2025.

\bibitem{hoskins1992process}
Josiah~C Hoskins and David~M Himmelblau.
\newblock Process control via artificial neural networks and reinforcement
  learning.
\newblock {\em Computers \& chemical engineering}, 16(4):241--251, 1992.

\bibitem{badgwell2018reinforcement}
Thomas~A Badgwell, Jay~H Lee, and Kuang-Hung Liu.
\newblock Reinforcement learning--overview of recent progress and implications
  for process control.
\newblock {\em Computer Aided Chemical Engineering}, 44:71--85, 2018.

\bibitem{nian2020review}
Rui Nian, Jinfeng Liu, and Biao Huang.
\newblock A review on reinforcement learning: Introduction and applications in
  industrial process control.
\newblock {\em Computers \& Chemical Engineering}, 139:106886, 2020.

\bibitem{pr13061791}
Venkata~Srikar Devarakonda, Wei Sun, Xun Tang, and Yuhe Tian.
\newblock Recent advances in reinforcement learning for chemical process
  control.
\newblock {\em Processes}, 13(6), 2025.

\bibitem{xiao2023modeling}
Ming Xiao, Cheng Hu, and Zhe Wu.
\newblock Modeling and predictive control of nonlinear processes using transfer
  learning method.
\newblock {\em AIChE Journal}, 69(7):e18076, 2023.

\bibitem{ArceMunoz.2024}
Samuel {Arce Munoz}, Jonathan Pershing, and John~D. Hedengren.
\newblock Physics-informed transfer learning for process control applications.
\newblock {\em Industrial {\&} Engineering Chemistry Research}, 2024.

\bibitem{Xiao.2024}
Ming Xiao, Keerthana Vellayappan, Pravin {P S}, Krishna Gudena, and Zhe Wu.
\newblock Optimization-based multi-source transfer learning for modeling of
  nonlinear processes.
\newblock {\em Chemical Engineering Science}, 295:120117, 2024.

\bibitem{morstyn2016control}
Thomas Morstyn, Branislav Hredzak, and Vassilios~G Agelidis.
\newblock Control strategies for microgrids with distributed energy storage
  systems: An overview.
\newblock {\em IEEE Transactions on Smart Grid}, 9(4):3652--3666, 2016.

\bibitem{otashu2021cooperative}
Joannah~I Otashu, Kyeongjun Seo, and Michael Baldea.
\newblock Cooperative optimal power flow with flexible chemical process loads.
\newblock {\em AIChE Journal}, 67(4):e17159, 2021.

\bibitem{dobbe2019toward}
Roel Dobbe, Oscar Sondermeijer, David Fridovich-Keil, Daniel Arnold, Duncan
  Callaway, and Claire Tomlin.
\newblock Toward distributed energy services: Decentralizing optimal power flow
  with machine learning.
\newblock {\em IEEE Transactions on Smart Grid}, 11(2):1296--1306, 2019.

\bibitem{lee2020federated}
Sangyoon Lee and Dae-Hyun Choi.
\newblock Federated reinforcement learning for energy management of multiple
  smart homes with distributed energy resources.
\newblock {\em IEEE Transactions on Industrial Informatics}, 18(1):488--497,
  2020.

\bibitem{li2023federated}
Yuanzheng Li, Shangyang He, Yang Li, Yang Shi, and Zhigang Zeng.
\newblock Federated multiagent deep reinforcement learning approach via
  physics-informed reward for multimicrogrid energy management.
\newblock {\em IEEE Transactions on Neural Networks and Learning Systems},
  2023.

\bibitem{joshi2023survey}
Aditya Joshi, Skieler Capezza, Ahmad Alhaji, and Mo-Yuen Chow.
\newblock Survey on ai and machine learning techniques for microgrid energy
  management systems.
\newblock {\em IEEE/CAA Journal of Automatica Sinica}, 10(7):1513--1529, 2023.

\bibitem{Allman.2022}
Andrew Allman and Qi~Zhang.
\newblock Distributed fairness-guided optimization for coordinated demand
  response in multi-stakeholder process networks.
\newblock {\em Computers {\&} Chemical Engineering}, 161:107777, 2022.

\bibitem{Li.2025}
Can Li.
\newblock Breaking data silos in drug discovery with federated learning.
\newblock {\em Nature Chemical Engineering}, 2(5):288--289, 2025.

\bibitem{choi2021deep}
Kukjin Choi, Jihun Yi, Changhwa Park, and Sungroh Yoon.
\newblock Deep learning for anomaly detection in time-series data: Review,
  analysis, and guidelines.
\newblock {\em IEEE access}, 9:120043--120065, 2021.

\bibitem{Hartung.2023}
Fabian Hartung, Billy~Joe Franks, Tobias Michels, Dennis Wagner, Philipp
  Liznerski, Steffen Reithermann, Sophie Fellenz, Fabian Jirasek, Maja Rudolph,
  Daniel Neider, Heike Leitte, Chen Song, Benjamin Kloepper, Stephan Mandt,
  Michael Bortz, Jakob Burger, Hans Hasse, and Marius Kloft.
\newblock Deep anomaly detection on tennessee eastman process data.
\newblock {\em Chemie Ingenieur Technik}, 95(7):1077--1082, 2023.

\bibitem{arunthavanathan2021analysis}
Rajeevan Arunthavanathan, Faisal Khan, Salim Ahmed, and Syed Imtiaz.
\newblock An analysis of process fault diagnosis methods from safety
  perspectives.
\newblock {\em Computers \& Chemical Engineering}, 145:107197, 2021.

\bibitem{Downs.1993}
J.~J. Downs and E.~F. Vogel.
\newblock A plant-wide industrial process control problem.
\newblock {\em Computers {\&} Chemical Engineering}, 17(3):245--255, 1993.

\bibitem{yang2019generative}
Xin Yang and Dajun Feng.
\newblock Generative adversarial network based anomaly detection on the
  benchmark tennessee eastman process.
\newblock In {\em 2019 5th International conference on control, automation and
  robotics (ICCAR)}, pages 644--648. IEEE, 2019.

\bibitem{lomov2021fault}
Ildar Lomov, Mark Lyubimov, Ilya Makarov, and Leonid~E Zhukov.
\newblock Fault detection in tennessee eastman process with temporal deep
  learning models.
\newblock {\em Journal of Industrial Information Integration}, 23:100216, 2021.

\bibitem{jia2023topology}
Mingwei Jia, Junhao Hu, Yi~Liu, Zengliang Gao, and Yuan Yao.
\newblock Topology-guided graph learning for process fault diagnosis.
\newblock {\em Industrial \& Engineering Chemistry Research}, 62(7):3238--3248,
  2023.

\bibitem{schoch2024deep}
Fabian Schoch, Pascal Graf, Tobias Schmieg, Carsten Wittenberg, Carsten
  Lanquillon, and Nicolaj~C Stache.
\newblock Deep anomaly detection with extended transformer-based model on
  tennessee eastman process dataset.
\newblock In {\em 2024 IEEE 19th International Conference on Computer Science
  and Information Technologies (CSIT)}, pages 1--4. IEEE, 2024.

\bibitem{Yan.2024}
Peng Yan, Ahmed Abdulkadir, Paul-Philipp Luley, Matthias Rosenthal, Gerrit~A.
  Schatte, Benjamin~F. Grewe, and Thilo Stadelmann.
\newblock A comprehensive survey of deep transfer learning for anomaly
  detection in industrial time series: Methods, applications, and directions.
\newblock {\em IEEE Access}, 12:3768--3789, 2024.

\bibitem{single2020knowledge}
Johannes~I Single, J{\"u}rgen Schmidt, and Jens Denecke.
\newblock Knowledge acquisition from chemical accident databases using an
  ontology-based method and natural language processing.
\newblock {\em Safety Science}, 129:104747, 2020.

\bibitem{feng2021application}
Xiayuan Feng, Yiyang Dai, Xu~Ji, Li~Zhou, and Yagu Dang.
\newblock Application of natural language processing in hazop reports.
\newblock {\em Process safety and environmental protection}, 155:41--48, 2021.

\bibitem{Niu.2024}
Yi~Niu, Yunxiao Fan, and Xing Ju.
\newblock Critical review on data-driven approaches for learning from
  accidents: Comparative analysis and future research.
\newblock {\em Safety Science}, 171:106381, 2024.

\bibitem{baltruvsaitis2018multimodal}
Tadas Baltru{\v{s}}aitis, Chaitanya Ahuja, and Louis-Philippe Morency.
\newblock Multimodal machine learning: A survey and taxonomy.
\newblock {\em IEEE transactions on pattern analysis and machine intelligence},
  41(2):423--443, 2018.

\bibitem{liang2024foundations}
Paul~Pu Liang, Amir Zadeh, and Louis-Philippe Morency.
\newblock Foundations \& trends in multimodal machine learning: Principles,
  challenges, and open questions.
\newblock {\em ACM Computing Surveys}, 56(10):1--42, 2024.

\bibitem{wu2024next}
Shengqiong Wu, Hao Fei, Leigang Qu, Wei Ji, and Tat-Seng Chua.
\newblock Next-gpt: Any-to-any multimodal llm.
\newblock In {\em Forty-first International Conference on Machine Learning},
  2024.

\bibitem{Scholtz.2024}
Ernst Scholtz, Alexandre Oudalov, and Iiro Harjunkoski.
\newblock Power systems of the future.
\newblock {\em Computers {\&} Chemical Engineering}, 180:108460, 2024.

\bibitem{Zhang.2025}
Xiangyu Zhang, Andrew Glaws, Alexandre Cortiella, Patrick Emami, and Ryan~N.
  King.
\newblock Deep generative models in energy system applications: Review,
  challenges, and future directions.
\newblock {\em Applied Energy}, 380:125059, 2025.

\bibitem{Tsay.2019}
Calvin Tsay and Michael Baldea.
\newblock 110th anniversary: Using data to bridge the time and length scales of
  process systems.
\newblock {\em Industrial {\&} Engineering Chemistry Research},
  58(36):16696--16708, 2019.

\bibitem{Coley.2017}
Connor~W. Coley, Regina Barzilay, William~H. Green, Tommi~S. Jaakkola, and
  Klavs~F. Jensen.
\newblock Convolutional embedding of attributed molecular graphs for physical
  property prediction.
\newblock {\em Journal of Chemical Information and Modeling}, 57(8):1757--1772,
  2017.

\bibitem{Rittig.2022}
Jan~G. Rittig, Qinghe Gao, Manuel Dahmen, Alexander Mitsos, and Artur~M.
  Schweidtmann.
\newblock Graph neural networks for the prediction of molecular
  structure--property relationships.
\newblock {\em Machine Learning and Hybrid Modelling for Reaction Engineering,
  Royal Society of Chemistry}, pages 159--181, 2022.

\bibitem{Reiser.2022}
Patrick Reiser, Marlen Neubert, Andr{\'e} Eberhard, Luca Torresi, Chen Zhou,
  Chen Shao, Houssam Metni, Clint {van Hoesel}, Henrik Schopmans, Timo Sommer,
  and Pascal Friederich.
\newblock Graph neural networks for materials science and chemistry.
\newblock {\em Communications Materials}, 3(1):93, 2022.

\bibitem{Felton_MLSAFT.2023}
Kobi~C Felton, Lukas Ra{\ss}pe-Lange, Jan~G Rittig, Kai Leonhard, Alexander
  Mitsos, Julian Meyer-Kirschner, Carsten Kn{\"o}sche, and Alexei~A Lapkin.
\newblock Ml-saft: a machine learning framework for pcp-saft parameter
  prediction.
\newblock {\em Chemical Engineering Journal}, page 151999, 2024.

\bibitem{Heid.2024}
Esther Heid, Kevin~P. Greenman, Yunsie Chung, Shih-Cheng Li, David~E. Graff,
  Florence~H. Vermeire, Haoyang Wu, William~H. Green, and Charles~J. McGill.
\newblock Chemprop: A machine learning package for chemical property
  prediction.
\newblock {\em Journal of Chemical Information and Modeling}, 64(1):9--17,
  2024.

\bibitem{chithrananda2020chemberta}
Seyone Chithrananda, Gabriel Grand, and Bharath Ramsundar.
\newblock Chemberta: large-scale self-supervised pretraining for molecular
  property prediction.
\newblock {\em arXiv preprint arXiv:2010.09885}, 2020.

\bibitem{chen2023generalizing}
Guzhong Chen, Zhen Song, Zhiwen Qi, and Kai Sundmacher.
\newblock Generalizing property prediction of ionic liquids from limited
  labeled data: a one-stop framework empowered by transfer learning.
\newblock {\em Digital Discovery}, 2(3):591--601, 2023.

\bibitem{winter2025understanding}
Benedikt Winter, Philipp Rehner, Timm Esper, Johannes Schilling, and Andr{\'e}
  Bardow.
\newblock Understanding the language of molecules: Predicting pure component
  parameters for the pc-saft equation of state from smiles.
\newblock {\em Digital Discovery}, 2025.

\bibitem{Chen.2021}
Shaoqi Chen, Dongyu Xue, Guohui Chuai, Qiang Yang, and Qi~Liu.
\newblock Fl-{QSAR}: a federated learning-based {QSAR} prototype for
  collaborative drug discovery.
\newblock {\em Bioinformatics (Oxford, England)}, 36(22-23):5492--5498, 2021.

\bibitem{Huang.2023}
Dong Huang, Xiucai Ye, Ying Zhang, and Tetsuya Sakurai.
\newblock Collaborative analysis for drug discovery by federated learning on
  non-iid data.
\newblock {\em Methods (San Diego, Calif.)}, 219:1--7, 2023.

\bibitem{Manu.2024}
Daniel Manu, Jingjing Yao, Wuji Liu, and Xiang Sun.
\newblock Graphganfed: A federated generative framework for graph-structured
  molecules towards efficient drug discovery.
\newblock {\em IEEE/ACM Transactions on Computational Biology and
  Bioinformatics}, 21(2):240--253, 2024.

\bibitem{Guo.2024}
Yan Guo, Yongqiang Gao, and Jiawei Song.
\newblock Molcfl: A personalized and privacy-preserving drug discovery
  framework based on generative clustered federated learning.
\newblock {\em Journal of biomedical informatics}, 157:104712, 2024.

\bibitem{Feng.2024}
Jinjia Feng, Zhen Wang, Zhewei Wei, Yaliang Li, Bolin Ding, and Hongteng Xu.
\newblock Federated heterogeneous contrastive distillation for molecular
  representation learning.
\newblock In {\em Serra, Spezzano (Ed.) 2024 -- Proceedings of the 33rd ACM},
  pages 1038--1048.

\bibitem{gomez2018automatic}
Rafael G{\'o}mez-Bombarelli, Jennifer~N Wei, David Duvenaud, Jos{\'e}~Miguel
  Hern{\'a}ndez-Lobato, Benjam{\'\i}n S{\'a}nchez-Lengeling, Dennis Sheberla,
  Jorge Aguilera-Iparraguirre, Timothy~D Hirzel, Ryan~P Adams, and Al{\'a}n
  Aspuru-Guzik.
\newblock Automatic chemical design using a data-driven continuous
  representation of molecules.
\newblock {\em ACS central science}, 4(2):268--276, 2018.

\bibitem{Bilodeau2022}
Camille Bilodeau, Wengong Jin, Tommi Jaakkola, Regina Barzilay, and Klavs~F
  Jensen.
\newblock Generative models for molecular discovery: Recent advances and
  challenges.
\newblock {\em Wiley Interdisciplinary Reviews: Computational Molecular
  Science}, 12(5):e1608, 2022.

\bibitem{decardi2024generative}
Benjamin Decardi-Nelson, Abdulelah~S Alshehri, Akshay Ajagekar, and Fengqi You.
\newblock Generative ai and process systems engineering: The next frontier.
\newblock {\em Computers \& Chemical Engineering}, page 108723, 2024.

\bibitem{zhang2023deep}
Jun Zhang, Qin Wang, Mario Eden, and Weifeng Shen.
\newblock A deep learning-based framework towards inverse green solvent design
  for extractive distillation with multi-index constraints.
\newblock {\em Computers \& Chemical Engineering}, 177:108335, 2023.

\bibitem{pirnay2025graphxform}
Jonathan Pirnay, Jan~G Rittig, Alexander~B Wolf, Martin Grohe, Jakob Burger,
  Alexander Mitsos, and Dominik~G Grimm.
\newblock Graphxform: graph transformer for computer-aided molecular design.
\newblock {\em Digital Discovery}, 4(4):1052--1065, 2025.

\bibitem{RittigRitzert_GraphMLFuel.2022}
Jan~G. Rittig, Martin Ritzert, Artur~M. Schweidtmann, Stefanie Winkler, Jana~M.
  Weber, Philipp Morsch, Karl~Alexander Heufer, Martin Grohe, Alexander Mitsos,
  and Manuel Dahmen.
\newblock Graph machine learning for design of high--octane fuels.
\newblock {\em AIChE Journal}, 69(4):e17971, 2023.

\bibitem{dos2021navigating}
Gabriel dos Passos~Gomes, Robert Pollice, and Al{\'a}n Aspuru-Guzik.
\newblock Navigating through the maze of homogeneous catalyst design with
  machine learning.
\newblock {\em Trends in Chemistry}, 3(2):96--110, 2021.

\bibitem{ishikawa2022heterogeneous}
Atsushi Ishikawa.
\newblock Heterogeneous catalyst design by generative adversarial network and
  first-principles based microkinetics.
\newblock {\em Scientific Reports}, 12(1):11657, 2022.

\bibitem{schilter2023designing}
Oliver Schilter, Alain Vaucher, Philippe Schwaller, and Teodoro Laino.
\newblock Designing catalysts with deep generative models and computational
  data. a case study for suzuki cross coupling reactions.
\newblock {\em Digital discovery}, 2(3):728--735, 2023.

\bibitem{SanchezMedina.2022}
Edgar~Ivan {Sanchez Medina}, Steffen Linke, Martin Stoll, and Kai Sundmacher.
\newblock Graph neural networks for the prediction of infinite dilution
  activity coefficients.
\newblock {\em Digital Discovery}, 1(3):216--225, 2022.

\bibitem{Rittig_GNNgammaIL.2022}
Jan~G. Rittig, Karim {Ben Hicham}, Artur~M. Schweidtmann, Manuel Dahmen, and
  Alexander Mitsos.
\newblock Graph neural networks for temperature-dependent activity coefficient
  prediction of solutes in ionic liquids.
\newblock {\em Computers and Chemical Engineering}, 171:108153, 2023.

\bibitem{Vermeire.2021}
Florence~H. Vermeire and William~H. Green.
\newblock Transfer learning for solvation free energies: From quantum chemistry
  to experiments.
\newblock {\em Chemical Engineering Journal}, 418:129307, August 2021.

\bibitem{Qin.2023}
Shiyi Qin, Shengli Jiang, Jianping Li, Prasanna Balaprakash, Reid~C. {van
  Lehn}, and Victor~M. Zavala.
\newblock Capturing molecular interactions in graph neural networks: a case
  study in multi-component phase equilibrium.
\newblock {\em Digital Discovery}, 2(1):138--151, 2023.

\bibitem{leenhouts2025pooling}
Roel~J Leenhouts, Nathan Morgan, Emad Al~Ibrahim, William~H Green, and
  Florence~H Vermeire.
\newblock Pooling solvent mixtures for solvation free energy predictions.
\newblock {\em Chemical Engineering Journal}, 513:162232, 2025.

\bibitem{Jirasek.2020}
Fabian Jirasek, Rodrigo A.~S. Alves, Julie Damay, Robert~A. Vandermeulen,
  Robert Bamler, Michael Bortz, Stephan Mandt, Marius Kloft, and Hans Hasse.
\newblock Machine learning in thermodynamics: Prediction of activity
  coefficients by matrix completion.
\newblock {\em The Journal of Physical Chemistry Letters}, 11(3):981--985,
  2020.

\bibitem{chen2021neural}
Guzhong Chen, Zhen Song, Zhiwen Qi, and Kai Sundmacher.
\newblock Neural recommender system for the activity coefficient prediction and
  unifac model extension of ionic liquid-solute systems.
\newblock {\em AIChE Journal}, 67(4):e17171, 2021.

\bibitem{Winter.2022}
Benedikt Winter, Clemens Winter, Johannes Schilling, and Andr{\'{e}} Bardow.
\newblock A smile is all you need: predicting limiting activity coefficients
  from {SMILES} with natural language processing.
\newblock {\em Digital Discovery}, 1(6):859--869, 2022.

\bibitem{abolhasani2023rise}
Milad Abolhasani and Eugenia Kumacheva.
\newblock The rise of self-driving labs in chemical and materials sciences.
\newblock {\em Nature Synthesis}, 2(6):483--492, 2023.

\bibitem{Tom.2024}
Gary Tom, Stefan~P. Schmid, Sterling~G. Baird, Yang Cao, Kourosh Darvish, Han
  Hao, Stanley Lo, Sergio Pablo-Garc{\'i}a, Ella~M. Rajaonson, Marta Skreta,
  Naruki Yoshikawa, Samantha Corapi, Gun~Deniz Akkoc, Felix Strieth-Kalthoff,
  Martin Seifrid, and Al{\'a}n Aspuru-Guzik.
\newblock Self-driving laboratories for chemistry and materials science.
\newblock {\em Chemical reviews}, 124(16):9633--9732, 2024.

\bibitem{Skilton.2015}
Ryan~A. Skilton, Richard~A. Bourne, Zacharias Amara, Raphael Horvath, Jing Jin,
  Michael~J. Scully, Emilia Streng, Samantha L.~Y. Tang, Peter~A. Summers,
  Jiawei Wang, Eduardo P{\'e}rez, Nigist Asfaw, Guilherme L.~P. Aydos, Jairton
  Dupont, Gurbuz Comak, Michael~W. George, and Martyn Poliakoff.
\newblock Remote-controlled experiments with cloud chemistry.
\newblock {\em Nature chemistry}, 7(1):1--5, 2015.

\bibitem{Fitzpatrick.2018}
Daniel~E. Fitzpatrick, Timoth{\'e} Maujean, Amanda~C. Evans, and Steven~V. Ley.
\newblock Across-the-world automated optimization and continuous-flow synthesis
  of pharmaceutical agents operating through a cloud-based server.
\newblock {\em Angewandte Chemie (International ed. in English)},
  57(46):15128--15132, 2018.

\bibitem{Bai.2024}
Jiaru Bai, Sebastian Mosbach, Connor~J. Taylor, Dogancan Karan, Kok~Foong Lee,
  Simon~D. Rihm, Jethro Akroyd, Alexei~A. Lapkin, and Markus Kraft.
\newblock A dynamic knowledge graph approach to distributed self-driving
  laboratories.
\newblock {\em Nature Communications}, 15(1):462, 2024.

\bibitem{Bommasani.2021}
Rishi Bommasani, Drew~A. Hudson, Ehsan Adeli, Russ Altman, Simran Arora, Sydney
  von Arx, Michael~S. Bernstein, Jeannette Bohg, Antoine Bosselut, Emma
  Brunskill, Erik Brynjolfsson, Shyamal Buch, Dallas Card, Rodrigo Castellon,
  Niladri Chatterji, Annie Chen, Kathleen Creel, Jared~Quincy Davis, Dora
  Demszky, Chris Donahue, Moussa Doumbouya, Esin Durmus, Stefano Ermon, John
  Etchemendy, Kawin Ethayarajh, Li~Fei-Fei, Chelsea Finn, Trevor Gale, Lauren
  Gillespie, Karan Goel, Noah Goodman, Shelby Grossman, Neel Guha, Tatsunori
  Hashimoto, Peter Henderson, John Hewitt, Daniel~E. Ho, Jenny Hong, Kyle Hsu,
  Jing Huang, Thomas Icard, Saahil Jain, Dan Jurafsky, Pratyusha Kalluri,
  Siddharth Karamcheti, Geoff Keeling, Fereshte Khani, Omar Khattab, Pang~Wei
  Koh, Mark Krass, Ranjay Krishna, Rohith Kuditipudi, Ananya Kumar, Faisal
  Ladhak, Mina Lee, Tony Lee, Jure Leskovec, Isabelle Levent, Xiang~Lisa Li,
  Xuechen Li, Tengyu Ma, Ali Malik, Christopher~D. Manning, Suvir Mirchandani,
  Eric Mitchell, Zanele Munyikwa, Suraj Nair, Avanika Narayan, Deepak
  Narayanan, Ben Newman, Allen Nie, Juan~Carlos Niebles, Hamed Nilforoshan,
  Julian Nyarko, Giray Ogut, Laurel Orr, Isabel Papadimitriou, Joon~Sung Park,
  Chris Piech, Eva Portelance, Christopher Potts, Aditi Raghunathan, Rob Reich,
  Hongyu Ren, Frieda Rong, Yusuf Roohani, Camilo Ruiz, Jack Ryan, Christopher
  R{\'e}, Dorsa Sadigh, Shiori Sagawa, Keshav Santhanam, Andy Shih, Krishnan
  Srinivasan, Alex Tamkin, Rohan Taori, Armin~W. Thomas, Florian Tram{\`e}r,
  Rose~E. Wang, William Wang, Bohan Wu, Jiajun Wu, Yuhuai Wu, Sang~Michael Xie,
  Michihiro Yasunaga, Jiaxuan You, Matei Zaharia, Michael Zhang, Tianyi Zhang,
  Xikun Zhang, Yuhui Zhang, Lucia Zheng, Kaitlyn Zhou, and Percy Liang.
\newblock On the opportunities and risks of foundation models.
\newblock {\em arXiv Preprint arXiv:2108.07258v3}, 2021.

\bibitem{Das.2024}
Abhimanyu Das, Weihao Kong, Rajat Sen, and Yichen Zhou.
\newblock {\em A decoder-only foundation model for time-series forecasting}.
\newblock 2024.

\bibitem{DecardiNelson.2024}
Benjamin Decardi-Nelson, Abdulelah~S. Alshehri, Akshay Ajagekar, and Fengqi
  You.
\newblock Generative ai and process systems engineering: The next frontier.
\newblock {\em Computers {\&} Chemical Engineering}, 187:108723, 2024.

\bibitem{Ren.2024}
Chao Ren, Han Yu, Hongyi Peng, Xiaoli Tang, Anran Li, Yulan Gao, Alysa~Ziying
  Tan, Bo~Zhao, Xiaoxiao Li, Zengxiang Li, and Qiang Yang.
\newblock {\em Advances and Open Challenges in Federated Learning with
  Foundation Models}.
\newblock 2024.

\bibitem{bran2023chemcrow}
Andres~M Bran, Sam Cox, Oliver Schilter, Carlo Baldassari, Andrew~D White, and
  Philippe Schwaller.
\newblock Chemcrow: Augmenting large-language models with chemistry tools.
\newblock {\em arXiv preprint arXiv:2304.05376}, 2023.

\bibitem{choi2025perspective}
Junyoung Choi, Gunwook Nam, Jaesik Choi, and Yousung Jung.
\newblock A perspective on foundation models in chemistry.
\newblock {\em JACS Au}, 5(4):1499--1518, 2025.

\bibitem{Huang.2022b}
Chao Huang, Shuqi Ke, Charles Kamhoua, Prasant Mohapatra, and Xin Liu.
\newblock Incentivizing data contribution in cross-silo federated learning.
\newblock {\em arXiv Preprint arXiv:2203.03885v2}, 2022.

\bibitem{Tu.2022}
Xuezhen Tu, Kun Zhu, Nguyen~Cong Luong, Dusit Niyato, Yang Zhang, and Juan Li.
\newblock Incentive mechanisms for federated learning: From economic and game
  theoretic perspective.
\newblock {\em IEEE Transactions on Cognitive Communications and Networking},
  8(3):1566--1593, 2022.

\bibitem{Huang.2022a}
Chao Huang, Jianwei Huang, and Xin Liu.
\newblock Cross-silo federated learning: Challenges and opportunities.
\newblock {\em arXiv Preprint arXiv:2206.12949v1}, 2022.

\bibitem{Huang.2024c}
Chao Huang, Justin Dachille, and Xin Liu.
\newblock When federated learning meets oligopoly competition: Stability and
  model differentiation.
\newblock {\em IEEE Internet of Things Journal}, 11(16):27409--27420, 2024.

\bibitem{Wu.2024}
Zhaoxuan Wu, Xinyi Xu, Rachael Hwee~Ling Sim, Yao Shu, Xiaoqiang Lin, Lucas
  Agussurja, Zhongxiang Dai, See-Kiong Ng, Chuan-Sheng Foo, Patrick Jaillet,
  Trong~Nghia Hoang, and Bryan Kian~Hsiang Low.
\newblock Chapter 15 - data valuation in federated learning.
\newblock In Lam~M. Nguyen, editor, {\em Federated Learning}, pages 281--296.
  {Elsevier Science {\&} Technology}, San Diego, 2024.

\bibitem{Sim.2024}
Rachael Hwee~Ling Sim, Sebastian~Shenghong Tay, Xinyi Xu, Yehong Zhang,
  Zhaoxuan Wu, Xiaoqiang Lin, See-Kiong Ng, Chuan-Sheng Foo, Patrick Jaillet,
  Trong~Nghia Hoang, and Bryan Kian~Hsiang Low.
\newblock Chapter 16 - incentives in federated learning.
\newblock In Lam~M. Nguyen, editor, {\em Federated Learning}, pages 299--309.
  {Elsevier Science {\&} Technology}, San Diego, 2024.

\bibitem{Huang.2024d}
Wenke Huang, Mang Ye, Zekun Shi, Guancheng Wan, He~Li, Bo~Du, and Qiang Yang.
\newblock Federated learning for generalization, robustness, fairness: A survey
  and benchmark.
\newblock {\em IEEE Transactions on Pattern Analysis and Machine Intelligence},
  PP:1--20, 2024.

\bibitem{Huang.2024a}
Chao Huang, Justin Dachille, and Xin Liu.
\newblock Technical report: Coopetition in heterogeneous cross-silo federated
  learning.

\bibitem{kruger2025publishing}
Fabian~P Kr{\"u}ger, Johan {\"O}stman, Lewis Mervin, Igor~V Tetko, and Ola
  Engkvist.
\newblock Publishing neural networks in drug discovery might compromise
  training data privacy.
\newblock {\em Journal of Cheminformatics}, 17(1):38, 2025.

\bibitem{bagdasaryan2020backdoor}
Eugene Bagdasaryan, Andreas Veit, Yiqing Hua, Deborah Estrin, and Vitaly
  Shmatikov.
\newblock How to backdoor federated learning.
\newblock In {\em International conference on artificial intelligence and
  statistics}, pages 2938--2948. PMLR, 2020.

\bibitem{Zhao.2024}
Joshua~C. Zhao, Saurabh Bagchi, Salman Avestimehr, Kevin~S. Chan, Somali
  Chaterji, Dimitris Dimitriadis, Jiacheng Li, Ninghui Li, Arash Nourian, and
  Holger~R. Roth.
\newblock Federated learning privacy: Attacks, defenses, applications, and
  policy landscape - a survey.
\newblock {\em arXiv Preprint arXiv:2405.03636v1}, 2024.

\bibitem{koay2023machine}
Abigail~MY Koay, Ryan K~L Ko, Hinne Hettema, and Kenneth Radke.
\newblock Machine learning in industrial control system (ics) security: current
  landscape, opportunities and challenges.
\newblock {\em Journal of Intelligent Information Systems}, 60(2):377--405,
  2023.

\bibitem{Cramer.2024}
Eike Cramer and Ji~Gao.
\newblock A black-box adversarial attack on demand side management.
\newblock {\em Computers {\&} Chemical Engineering}, 186:108681, 2024.

\bibitem{Heid.2023}
Esther Heid, Charles~J. McGill, Florence~H. Vermeire, and William~H. Green.
\newblock Characterizing uncertainty in machine learning for chemistry.
\newblock {\em Journal of Chemical Information and Modeling},
  63(13):4012--4029, 2023.

\bibitem{linsner2021approaches}
Florian Linsner, Linara Adilova, Sina D{\"a}ubener, Michael Kamp, and Asja
  Fischer.
\newblock Approaches to uncertainty quantification in federated deep learning.
\newblock In {\em Joint European Conference on Machine Learning and Knowledge
  Discovery in Databases}, pages 128--145. Springer, 2021.

\bibitem{zhang2023uncertainty}
Yuwei Zhang, Tong Xia, Abhirup Ghosh, and Cecilia Mascolo.
\newblock Uncertainty quantification in federated learning for heterogeneous
  health data.
\newblock In {\em International Workshop on Federated Learning for Distributed
  Data Mining}, 2023.

\bibitem{tuor2021overcoming}
Tiffany Tuor, Shiqiang Wang, Bong~Jun Ko, Changchang Liu, and Kin~K Leung.
\newblock Overcoming noisy and irrelevant data in federated learning.
\newblock In {\em 2020 25th International Conference on Pattern Recognition
  (ICPR)}, pages 5020--5027. IEEE, 2021.

\bibitem{Fang_2022_CVPR}
Xiuwen Fang and Mang Ye.
\newblock Robust federated learning with noisy and heterogeneous clients.
\newblock In {\em Proceedings of the IEEE/CVF Conference on Computer Vision and
  Pattern Recognition (CVPR)}, pages 10072--10081, June 2022.

\bibitem{yang2022robust}
Seunghan Yang, Hyoungseob Park, Junyoung Byun, and Changick Kim.
\newblock Robust federated learning with noisy labels.
\newblock {\em IEEE Intelligent Systems}, 37(2):35--43, 2022.

\bibitem{tsouvalas2024labeling}
Vasileios Tsouvalas, Aaqib Saeed, Tanir Ozcelebi, and Nirvana Meratnia.
\newblock Labeling chaos to learning harmony: Federated learning with noisy
  labels.
\newblock {\em ACM Transactions on Intelligent Systems and Technology},
  15(2):1--26, 2024.

\bibitem{Woisetschläger.2024a}
Herbert Woisetschl{\"a}ger, Alexander Erben, Bill Marino, Shiqiang Wang,
  Nicholas~D. Lane, Ruben Mayer, and Hans-Arno Jacobsen.
\newblock Federated learning priorities under the european union artificial
  intelligence act.
\newblock {\em arXiv Preprint arXiv:2402.05968v1}, 2024.

\bibitem{Chik.2024}
Warren Chik and Florian Gamper.
\newblock Chapter 21 - ethical considerations and legal issues relating to
  federated learning.
\newblock In Lam~M. Nguyen, editor, {\em Federated Learning}, pages 369--391.
  {Elsevier Science {\&} Technology}, San Diego, 2024.

\bibitem{aibs.2024}
Aisb consortium: https://www.apheris.com/industries/aisb.

\bibitem{Liu.2024b}
Rui Liu, Pengwei Xing, Zichao Deng, Anran Li, Cuntai Guan, and Han Yu.
\newblock Federated graph neural networks: Overview, techniques, and
  challenges.
\newblock {\em IEEE transactions on neural networks and learning systems}, PP,
  2024.

\bibitem{Kruger.2024}
Fabian~P. Kr{\"u}ger, Johan {\"O}stman, Lewis Mervin, Igor~V. Tetko, and Ola
  Engkvist.
\newblock Publishing neural networks in drug discovery might compromise
  training data privacy.

\bibitem{Winter.2023}
Benedikt Winter, Clemens Winter, Timm Esper, Johannes Schilling, and Andr{\'e}
  Bardow.
\newblock Spt-nrtl: A physics-guided machine learning model to predict
  thermodynamically consistent activity coefficients.
\newblock {\em Fluid Phase Equilibria}, 568:113731, 2023.

\bibitem{specht2024hanna}
Thomas Specht, Mayank Nagda, Sophie Fellenz, Stephan Mandt, Hans Hasse, and
  Fabian Jirasek.
\newblock Hanna: hard-constraint neural network for consistent activity
  coefficient prediction.
\newblock {\em Chemical Science}, 15(47):19777--19786, 2024.

\bibitem{ottaiano2024machine}
Gabriel~Y Ottaiano and Tiago~D Martins.
\newblock Machine learning models for vapor-liquid equilibrium of binary
  mixtures: State of the art and future opportunities.
\newblock {\em Chemical Engineering Research and Design}, 2024.

\bibitem{Felton.2022}
K.~C. Felton, H.~Ben-Safar, and A.~A. Alexei.
\newblock {DeepGamma}: A deep learning model for activity coefficient
  prediction.
\newblock In {\em 1st Annual AAAI Workshop on AI to Accelerate Science and
  Engineering (AI2ASE)}, 2022.

\bibitem{SanchezMedina.2023}
Edgar~Ivan {Sanchez Medina}, Steffen Linke, Martin Stoll, and Kai Sundmacher.
\newblock Gibbs--helmholtz graph neural network: capturing the temperature
  dependency of activity coefficients at infinite dilution.
\newblock {\em Digital Discovery}, 2:781--798, 2023.

\bibitem{Rittig.2023a}
Jan~G. Rittig, Karim {Ben Hicham}, Artur~M. Schweidtmann, Manuel Dahmen, and
  Alexander Mitsos.
\newblock Graph neural networks for temperature-dependent activity coefficient
  prediction of solutes in ionic liquids.
\newblock {\em Computers {\&} Chemical Engineering}, 171:108153, 2023.

\bibitem{Rittig.2023b}
Jan~G. Rittig, Kobi~C. Felton, Alexei~A. Lapkin, and Alexander Mitsos.
\newblock Gibbs--duhem-informed neural networks for binary activity coefficient
  prediction.
\newblock {\em Digital Discovery}, 2(6):1752--1767, 2023.

\bibitem{Gilmer.2017}
Justin Gilmer, Samuel~S. Schoenholz, Patrick~F. Riley, Oriol Vinyals, and
  George~E. Dahl.
\newblock Neural message passing for quantum chemistry.
\newblock {\em International Conference on Machine Learning}, pages 1263--1272,
  2017.

\bibitem{Brouwer.2019}
Thomas Brouwer and Boelo Schuur.
\newblock Model performances evaluated for infinite dilution activity
  coefficients prediction at 298.15 k.
\newblock {\em Industrial {\&} Engineering Chemistry Research},
  58(20):8903--8914, 2019.

\bibitem{Arivazhagan.2019}
Manoj~Ghuhan Arivazhagan, Vinay Aggarwal, Aaditya~Kumar Singh, and Sunav
  Choudhary.
\newblock Federated learning with personalization layers.
\newblock {\em arXiv Preprint arXiv:1912.00818v1}, 2019.

\bibitem{Pillutla.2022}
Krishna Pillutla, Kshitiz Malik, Abdel-Rahman Mohamed, Mike Rabbat, Maziar
  Sanjabi, and Lin Xiao.
\newblock Federated learning with partial model personalization.
\newblock {\em International Conference on Machine Learning}, pages
  17716--17758, 2022.

\bibitem{Fey.2019}
Matthias Fey and Jan~Eric Lenssen.
\newblock Fast graph representation learning with pytorch geometric.
\newblock {\em arXiv Preprint arXiv:1903.02428v3}, 2019.

\bibitem{rdkit}
Greg Landrum.
\newblock Rdkit: Open-source cheminformatics.

\bibitem{Beutel.2020}
Daniel~J. Beutel, Taner Topal, Akhil Mathur, Xinchi Qiu, Javier
  Fernandez-Marques, Yan Gao, Lorenzo Sani, Kwing~Hei Li, Titouan Parcollet,
  Pedro Porto~Buarque de~Gusm{\~a}o, and Nicholas~D. Lane.
\newblock Flower: A friendly federated learning research framework.
\newblock {\em arXiv Preprint arXiv:2007.14390v5}, 2020.

\bibitem{Wu.2024b}
Zhe Wu, Panagiotis~D. Christofides, Wanlu Wu, Yujia Wang, Fahim Abdullah, Aisha
  Alnajdi, and Yash Kadakia.
\newblock A tutorial review of machine learning-based model predictive control
  methods.
\newblock {\em Reviews in Chemical Engineering}, 2024.

\bibitem{niu2022deep}
Kaicheng Niu, Mi~Zhou, Chaouki~T Abdallah, and Mohammad Hayajneh.
\newblock Deep transfer learning for system identification using long
  short-term memory neural networks.
\newblock {\em arXiv preprint arXiv:2204.03125}, 2022.

\bibitem{Schulze.2022a}
Jan~C. Schulze, Danimir~T. Doncevic, and Alexander Mitsos.
\newblock Identification of mimo wiener-type koopman models for data-driven
  model reduction using deep learning.
\newblock {\em Computers {\&} Chemical Engineering}, 161:107781, 2022.

\bibitem{Jacobsen.1991}
Elling~W. Jacobsen and Sigurd Skogestad.
\newblock Multiple steady states in ideal two--product distillation.
\newblock {\em AIChE Journal}, 37(4):499--511, 1991.

\bibitem{Pearson.2000}
Ronald~K. Pearson and Martin Pottmann.
\newblock Gray-box identification of block-oriented nonlinear models.
\newblock {\em Journal of Process Control}, 10(4):301--315, 2000.

\bibitem{TensorFlowDevelopers.2024}
{TensorFlow Developers}.
\newblock Tensorflow, 2024.

\bibitem{Hart.2011}
William~E. Hart, Jean-Paul Watson, and David~L. Woodruff.
\newblock Pyomo: modeling and solving mathematical programs in python.
\newblock {\em Mathematical Programming Computation}, 3(3):219--260, 2011.

\bibitem{Bynum.2021}
Michael~L. Bynum, Gabriel~A. Hackebeil, William~E. Hart, Carl~D. Laird,
  Bethany~L. Nicholson, John~D. Siirola, Jean-Paul Watson, and David~L.
  Woodruff.
\newblock {\em Pyomo - optimization modeling in Python}, volume volume 67 of
  {\em Springer eBook Collection}.
\newblock Springer, Cham, third edition edition, 2021.

\bibitem{Andersson.2019}
Joel A.~E. Andersson, Joris Gillis, Greg Horn, James~B. Rawlings, and Moritz
  Diehl.
\newblock Casadi: a software framework for nonlinear optimization and optimal
  control.
\newblock {\em Mathematical Programming Computation}, 11(1):1--36, 2019.

\bibitem{Velioglu.2025}
Mehmet Velioglu, Song Zhai, Sophia Rupprecht, Alexander Mitsos, Andreas Jupke,
  and Manuel Dahmen.
\newblock Physics-informed neural networks for dynamic process operations with
  limited physical knowledge and data.
\newblock {\em Computers {\&} Chemical Engineering}, 192:108899, 2025.

\bibitem{Chen.2024}
Hao Chen, Gonzalo E.~Constante Flores, and Can Li.
\newblock Physics-informed neural networks with hard linear equality
  constraints.
\newblock {\em Computers {\&} Chemical Engineering}, 189:108764, 2024.

\end{thebibliography}

\end{document}